%% file: edges.tex
\documentclass[10pt,journal,letterpaper,twoside,compsoc]{IEEEtran}

\include{preamble}

\begin{document}

\title{Fast Edge Detection Using Structured Forests}
\author{
  Piotr Doll\'ar and C. Lawrence Zitnick\\
  Microsoft Research\\
  {\tt\small\{pdollar,larryz\}@microsoft.com}
}

\IEEEcompsoctitleabstractindextext{\begin{abstract}
Edge detection is a critical component of many vision systems, including object detectors and image segmentation algorithms. Patches of edges exhibit well-known forms of local structure, such as straight lines or T-junctions. In this paper we take advantage of the structure present in local image patches to learn both an accurate and computationally efficient edge detector. We formulate the problem of predicting local edge masks in a structured learning framework applied to random decision forests. Our novel approach to learning decision trees robustly maps the structured labels to a discrete space on which standard information gain measures may be evaluated. The result is an approach that obtains realtime performance that is orders of magnitude faster than many competing state-of-the-art approaches, while also achieving state-of-the-art edge detection results on the BSDS500 Segmentation dataset and NYU Depth dataset. Finally, we show the potential of our approach as a general purpose edge detector by showing our learned edge models generalize well across datasets.
\end{abstract}}\maketitle

\section{Introduction}

Edge detection has remained a fundamental task in computer vision since the early 1970's \cite{Fram1975IEEE,Duda1973BOOK,Robinson1977OE}. The detection of edges is a critical preprocessing step for a variety of tasks, including object recognition \cite{UllmanPAMI91,Ferrari2008PAMI}, segmentation \cite{Malik2001IJCV,Arbelaez2011PAMI}, and active contours \cite{Kass1988IJCV}. Traditional approaches to edge detection use a variety of methods for computing color gradients followed by non-maximal suppression \cite{Canny1986PAMI,Freeman91PAMI,Ziou1998PR}. Unfortunately, many visually salient edges do not correspond to color gradients, such as texture edges \cite{Martin2004PAMI} and illusory contours~\cite{Ren2005ICCV}. State-of-the-art edge detectors \cite{Arbelaez2011PAMI,Ren2012NIPS,Lim2013CVPR,Gupta2013CVPR} use multiple features as input, including brightness, color, texture and depth gradients computed over multiple scales.

Since visually salient edges correspond to a variety of visual phenomena, finding a unified approach to edge detection is difficult. Motivated by this observation several recent papers have explored the use of learning techniques for edge detection \cite{Dollar2006CVPR,Zheng07CVPR,Lim2013CVPR,Kivinen2014AISTATS}. These approaches take an image patch and compute the likelihood that the center pixel contains an edge. Optionally, the independent edge predictions may then be combined using global reasoning \cite{Arbelaez2011PAMI,Ren2012NIPS,Zheng07CVPR,Arbelaez2014CVPR}.

Edges in a local patch are highly interdependent \cite{Lim2013CVPR}. They often contain well-known patterns, such as straight lines, parallel lines, T-junctions or Y-junctions \cite{Ren2006ECCV,Lim2013CVPR}.  Recently, a family of learning approaches called {\it structured learning} \cite{Nowozin2011foundations} has been applied to problems exhibiting similar characteristics. For instance, \cite{Kontschieder2011ICCV} applies structured learning to the problem of semantic image labeling for which local image labels are also highly interdependent.

\begin{figure} \center
  \includegraphics[width=.475\textwidth]{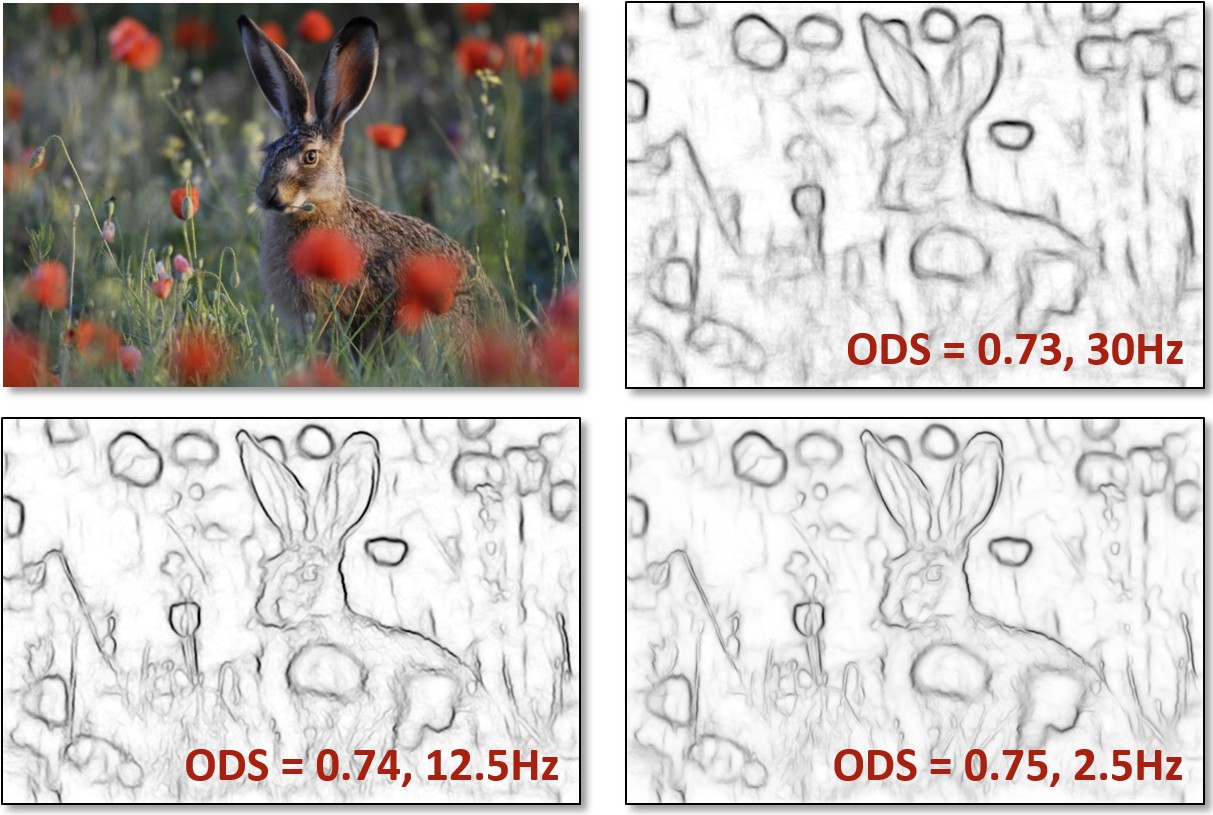}
\Caption{ Edge detection results using three versions of our Structured Edge (SE) detector demonstrating tradeoffs in accuracy vs. runtime.  We obtain realtime performance while simultaneously achieving state-of-the-art results. ODS numbers were computed on BSDS \cite{Arbelaez2011PAMI} on which the popular gPb detector \cite{Arbelaez2011PAMI} achieves a score of $.73$. The variants shown include SE, SE+SH, and SE+MS+SH, see \secref{sec:edges} for details. }
 \label{fig:teaser}\vspace{-1mm}
\end{figure}

In this paper we propose a generalized structured learning approach that we apply to edge detection. This approach allows us to take advantage of the inherent structure in edge patches, while being surprisingly computationally efficient. We can compute edge maps in realtime, which is orders of magnitude faster than competing state-of-the-art approaches. A random forest framework is used to capture the structured information \cite{Kontschieder2011ICCV}. We formulate the problem of edge detection as predicting local segmentation masks given input image patches.  Our novel approach to learning decision trees uses structured labels to determine the splitting function at each branch in the tree. The structured labels are robustly mapped to a discrete space on which standard information gain measures may be evaluated. Each forest predicts a patch of edge pixel labels that are aggregated across the image to compute our final edge map, see \figref{fig:teaser}. Since the aggregated edge maps may be diffuse, the edge maps may optionally be sharpened using local color and depth cues. We show state-of-the-art results on both the BSDS500 \cite{Arbelaez2011PAMI} and the NYU Depth dataset \cite{Silberman2012ECCV}. We demonstrate the potential of our approach as a general purpose edge detector by showing the strong cross dataset generalization of our learned edge models.

\subsection{Related work}

We now discuss related work in edge detection and structured learning. An earlier version of this work appeared in \cite{Dollar2013ICCV}.

\textbf{Edge detection:} Numerous papers have been written on edge detection over the past 50 years. Early work~\cite{Fram1975IEEE,Duda1973BOOK,Canny1986PAMI,Perona1990PAMI,Freeman91PAMI} focused on the detection of intensity or color gradients. The popular Canny detector \cite{Canny1986PAMI} finds the peak gradient orthogonal to edge direction. An evaluation of various low-level edge detectors can be found in \cite{Bowyer2001CVIU} and an overview in \cite{Ziou1998PR}. More recent work \cite{Martin2004PAMI,Mairal2008ECCV,Kokkinos2010ECCV,Arbelaez2011PAMI,Widynski2012ECCV,Leordeanu2014PAMI} explores edge detection under more challenging conditions.

Several techniques have explored the use of learning for edge detection \cite{Dollar2006CVPR,Zheng07CVPR,Mairal2008ECCV,Ren2012NIPS,Lim2013CVPR,Kivinen2014AISTATS}. Doll\'ar~\etal \cite{Dollar2006CVPR} used a boosted classifier to independently label each pixel using its surrounding image patch as input. Zheng~\etal \cite{Zheng07CVPR} combine low, mid, and high-level cues and show improved results for object-specific edge detection. Ren and Bo \cite{Ren2012NIPS} improved the result of \cite{Arbelaez2011PAMI} by computing gradients across learned sparse codes of patch gradients. While \cite{Ren2012NIPS} achieved good results, their approach further increased the high computational cost of \cite{Arbelaez2011PAMI}. Catanzaro~\etal \cite{Catanzaro2009ICCV} improve the runtime of \cite{Arbelaez2011PAMI} using parallel algorithms. Recently, Kivinen \etal~\cite{Kivinen2014AISTATS} applied deep networks to edge detection achieving competitive results.

Finally, Lim~\etal \cite{Lim2013CVPR} propose an edge detection approach that classifies edge patches into {\it sketch tokens} using random forest classifiers, that, like in our work, attempt to capture local edge structure. Sketch tokens bear resemblance to earlier work on {\it shapemes} \cite{Ren2006ECCV} but are computed directly from color image patches rather than from pre-computed edge maps. The result is an efficient approach for detecting edges that also shows promising results for object detection. In contrast to previous work, we do not require the use of pre-defined classes of edge patches. This allows us to learn more subtle variations in edge structure and leads to a more accurate and efficient algorithm.

\textbf{Structured learning:} Structured learning addresses the problem of learning a mapping where the input or output space may be arbitrarily complex representing strings, sequences, graphs, object pose, bounding boxes \etc. \cite{Tsochantaridis2004ICML,Taskar2005ICML,Blaschko2008ECCV}. We refer readers to \cite{Nowozin2011foundations} for a comprehensive survey.

Our structured random forests differ from these works in several respects. First, we assume that the output space is structured but operate on a standard input space. Second, by default our model can only output examples observed during training, which implicitly assumes the existence of a set of representative samples (this shortcoming can be ameliorated with custom ensemble models). On the other hand, typical structured predictors learn parameters to a scoring function and at inference perform an optimization to obtain predictions \cite{Tsochantaridis2004ICML,Nowozin2011foundations}. This requires defining a scoring function and an efficient (possibly approximate) inference procedure. In contrast, inference using our structured random forest is straightforward, general and fast (same as for standard random forests).

Finally, our work was inspired by recent work from Kontschieder \etal \cite{Kontschieder2011ICCV} on learning random forests for structured class labels for the specific case where the output labels represent a semantic image labeling for an image patch. The key observation made by Kontschieder \etal is that given a color image patch, the leaf node reached in a tree is independent of the structured semantic labels, and any type of output can be stored at each leaf. Building on this, we propose a general learning framework for structured output forests that can be used with a broad class of output spaces. We apply our framework to learning an accurate and fast edge detector.

\section{Random Decision Forests}\label{sec:forests}

We begin with a review of random decision forests \cite{Breiman1984BOOK,Breiman01ML,Geurts2006ML}. Throughout our presentation we adopt the notation and terminology of the extensive recent survey by Criminisi \etal \cite{Criminisi12FT}, somewhat simplified for ease of presentation. The notation in \cite{Criminisi12FT} is sufficiently general to support our extension to random forests with structured outputs.

A decision tree $f_t(x)$ classifies a sample $x \in \X$ by recursively branching left or right down the tree until a leaf node is reached. Specifically, each node $j$ in the tree is associated with a binary \textit{split function}: \eqn{\normalsize}{h(x,\theta_j) \in \{0,1\}}with parameters $\theta_j$. If $h(x,\theta_j)=0$ node $j$ sends $x$ left, otherwise right, with the process terminating at a leaf node. The output of the tree on an input $x$ is the prediction stored at the leaf reached by $x$, which may be a target label $y \in \Y$ or a distribution over the labels $\Y$.

While the split function $h(x,\theta)$ may be arbitrarily complex, a common choice is a `stump' where a single feature dimension of $x$ is compared to a threshold. Specifically, $\theta=(k,\tau)$ and $h(x,\theta) = [x(k) < \tau]$, where $[\cdot]$ denotes the indicator function. Another popular choice is $\theta=(k_1,k_2,\tau)$ and $h(x,\theta) = [x(k_1)-x(k_2)< \tau]$. Both are computationally efficient and effective in practice \cite{Criminisi12FT}.

A decision forest is an ensemble of $T$ independent trees $f_t$. Given a sample $x$, the predictions $f_t(x)$ from the set of trees are combined using an \textit{ensemble model} into a single output. Choice of ensemble model is problem specific and depends on $\Y$, common choices include majority voting for classification and averaging for regression, although more sophisticated ensemble models may be employed \cite{Criminisi12FT}.

Observe that arbitrary information may be stored at the leaves of a decision tree. The leaf node reached by the tree depends only on the input $x$, and while predictions of multiple trees must be merged in some useful way (the ensemble model), any type of output $y$ can be stored at each leaf. This allows use of complex output spaces $\Y$, including structured outputs as observed by Kontschieder \etal \cite{Kontschieder2011ICCV}.

While prediction is straightforward, training random decision forests with structured $\Y$ is more challenging. We review the standard learning procedure next and describe our generalization to learning with structured outputs in \secref{sec:structured}.

\subsection{Training Decision Trees}\label{sec:forests:training}

Each tree is trained independently in a recursive manner. For a given node $j$ and training set $\XY_j \subset \X \times \Y$, the goal is to find parameters $\theta_j$ of the split function $h(x,\theta_j)$ that result in a `good' split of the data. This requires defining an \textit{information gain criterion} of the form: \eqn{\normalsize}{I_j = I(\XY_j,\XY_j^L,\XY_j^R)\label{eqn:infogain0}}where $\XY_j^L=\{ (x,y)\in \XY_j | h(x,\theta_j)=0 \}$, $\XY_j^R = \XY_j \backslash \XY_j^L$. Splitting parameters $\theta_j$ are chosen to maximize the information gain $I_j$; training then proceeds recursively on the left node with data $\XY_j^L$ and similarly for the right node. Training stops when a maximum depth is reached or if information gain or training set size fall below fixed thresholds.

For multiclass classification ($\Y \subset \mathbb{Z}$) the standard definition of information gain can be used:
\eqn{\normalsize}{
  I_j=H(\XY_j)-\sum_{k\in\{L,R\}} \frac{|S_j^k|}{|S_j|}H(S_j^k)
\label{eqn:infogain}}%
where $H(\XY)=-\sum_y p_y \log(p_y)$ denotes the Shannon entropy and $p_y$ is the fraction of elements in $\XY$ with label $y$. Alternatively the Gini impurity $H(\XY)=\sum_y p_y (1-p_y)$ has also been used in conjunction with \eqnref{eqn:infogain} \cite{Breiman1984BOOK}.

For regression, entropy and information gain can be extended to continuous variables \cite{Criminisi12FT}. Alternatively, a common approach for single-variate regression ($\Y = \R$) is to minimize the variance of labels at the leaves \cite{Breiman1984BOOK}. If we write the variance as $H(S)=\frac{1}{|\XY|}\sum_y (y-\mu)^2$ where $\mu = \frac{1}{|\XY|}\sum_y y$, then substituting $H$ for entropy in \eqnref{eqn:infogain} leads to the standard criterion for single-variate regression.

Can we define a more general \textit{information gain} criterion for \eqnref{eqn:infogain0} that generalizes well for arbitrary output spaces $\Y$? Surprisingly yes, given mild additional assumptions about $\Y$. Before going into detail in \secref{sec:structured}, we discuss the key role that randomness plays in the training of decision forests next.

\subsection{Randomness and Optimality}\label{sec:forests:randomness}

Individual decision trees exhibit high variance and tend to overfit \cite{Ho1998PAMI,Breiman1984BOOK,Breiman01ML,Geurts2006ML}. Decision forests ameliorate this by training multiple de-correlated trees and combining their output. A crucial component of the training procedure is therefore to achieve a sufficient diversity of trees.

Diversity of trees can be obtained either by randomly subsampling the data used to train each \textit{tree} \cite{Breiman1984BOOK} or randomly subsampling the features and splits used to train each \textit{node} \cite{Ho1998PAMI}. Injecting randomness at the level of nodes tends to produce higher accuracy models \cite{Geurts2006ML} and has proven more popular \cite{Criminisi12FT}. Specifically, when optimizing \eqnref{eqn:infogain0}, only a small set of possible $\theta_j$ are sampled and tested when choosing the optimal split. \Eg, for stumps where $\theta=(k,\tau)$ and $h(x,\theta) = [x(k) < \tau]$, \cite{Geurts2006ML} advocates sampling $\sqrt{d}$ features where $\X = \R^d$ and a single threshold $\tau$ per feature.

In effect, accuracy of individual trees is sacrificed in favor of a high diversity ensemble \cite{Geurts2006ML}. Leveraging similar intuition allows us to introduce an approximate information gain criterion for structured labels, described next, and leads to our generalized structured forest formulation.

\section{Structured Random Forests}\label{sec:structured}

In this section we extend random decision forests to general structured output spaces $\Y$. Of particular interest for computer vision is the case where $x\in\X$ represents an image patch and $y\in\Y$ encodes the corresponding local image annotation (\eg, a segmentation mask or set of semantic image labels). However, we keep our derivation general.

Training random forests with structured labels poses two main challenges. First, structured output spaces are often high dimensional and complex. Thus scoring numerous candidate splits directly over structured labels may be prohibitively expensive. Second, and more critically, information gain over structured labels may not be well defined.

We use the observation that even \textit{approximate} measures of information gain suffice to train effective random forest classifiers \cite{Geurts2006ML,Kontschieder2011ICCV}. `Optimal' splits are not necessary or even desired, see \secref{sec:forests:randomness}. Our core idea is to map all the structured labels $y\in\Y$ at a given node into a discrete set of labels $c \in \C$, where $\C=\{1,\ldots,k\}$, such that \textit{similar} structured labels $y$ are assigned to the same discrete label $c$.

Given the discrete labels $\C$, information gain calculated directly and efficiently over $\C$ can serve as a proxy for the information gain over the structured labels $\Y$. As a result at each node we can leverage existing random forest training procedures to learn structured random forests effectively.

Our approach to calculating information gain relies on measuring similarity over $\Y$. However, for many structured output spaces, including those used for edge detection, computing similarity over $\Y$ is not well defined. Instead, we define a mapping of $\Y$ to an \textit{intermediate} space $\Z$ in which distance is easily measured. We therefore utilize a broadly applicable two-stage approach of first mapping $\Y\rightarrow\Z$ followed by a straightforward mapping of $\Z\rightarrow\C$.

We describe the proposed approach in more detail next and return to its application to edge detection in \secref{sec:edges}.

\begin{figure*} \center
  \includegraphics[width=.9\textwidth]{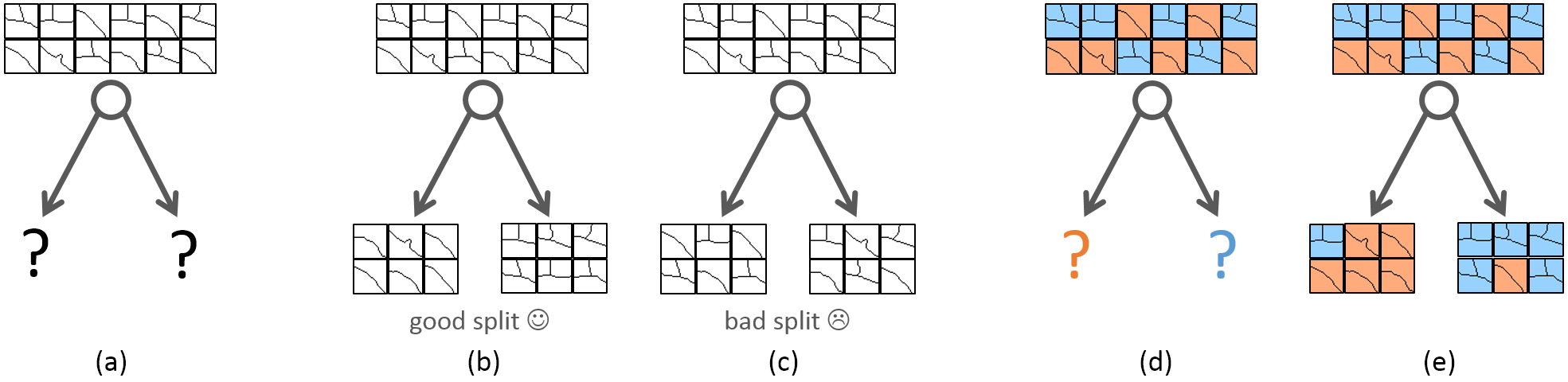}\vspace{-2mm}
\Caption{Illustration of the decision tree node splits: (a) Given a set of structured labels such as segments, a splitting function must be determined. Intuitively a good split (b) groups similar segments, whereas a bad split (c) does not. In practice we cluster the structured labels into two classes (d). Given the class labels, a standard splitting criterion, such as Gini impurity, may be used (e).}
\label{fig:split}\vspace{-2mm}
\end{figure*}

\subsection{Intermediate Mapping $\map$}\label{sec:structured:mapping}

Our key assumption is that for many structured output spaces, including for structured learning of edge detection, we can define a mapping of the form: \eqn{\normalsize}{\map : \Y \rightarrow \Z\label{eqn:map}}such that we can approximate dissimilarity of $y\in\Y$ by computing Euclidean distance in $\Z$. For example, as we describe in detail in \secref{sec:edges}, for edge detection the labels $y\in\Y$ are $16 \times 16$ segmentation masks and we define $z=\map(y)$ to be a long binary vector that encodes whether every pair of pixels in $y$ belong to the same or different segments. Distance is easily measured in the resulting space $\Z$.

$\Z$ may be high dimensional which presents a challenge computationally. For example, for edge detection there are $\binom{16 \cdot 16}{2} = 32640$ unique pixel pairs in a $16 \times 16$ segmentation mask, so computing $z$ for every $y$ would be expensive. However, as only an approximate distance measure is necessary, the dimensionality of $\Z$ can be reduced.

In order to reduce dimensionality, we sample $m$ dimensions of $\Z$, resulting in a reduced mapping $\map_\phi : \Y \rightarrow \Z$ parametrized by $\phi$. During training, a distinct mapping $\map_\phi$ is randomly generated and applied to training labels $\Y_j$ at each node $j$. This serves two purposes. First, $\map_\phi$ can be considerably faster to compute than $\map$. Second, sampling $\Z$ injects additional randomness into the learning process and helps ensure a sufficient diversity of trees, see \secref{sec:forests:randomness}.

Finally, Principal Component Analysis (PCA) \cite{Joliffe1986BOOK} can be used to further reduce the dimensionality of $\Z$. PCA denoises $\Z$ while approximately preserving Euclidean distance. In practice, we use $\map_\phi$ with $m=256$ dimensions followed by a PCA projection to at most 5 dimensions.

\subsection{Information Gain Criterion}\label{sec:structured:information}

Given the mapping $\map_\phi:\Y\rightarrow\Z$, a number of choices for the information gain criterion are possible. For discrete $\Z$ multi-variate joint entropy could be computed directly. Kontschieder \etal \cite{Kontschieder2011ICCV} proposed such an approach, but due to its complexity of $O(|\Z|^m)$, were limited to using $m\le2$. Our experiments indicate $m\ge64$ is necessary to accurately capture similarities between elements in $\Z$. Alternatively, given continuous $\Z$, variance or a continuous formulation of entropy \cite{Criminisi12FT} can be used to define information gain. In this work we propose a simpler, extremely efficient approach.

We map a set of structured labels $y \in \Y$ into a discrete set of labels $c \in \C$, where $\C=\{1,\ldots,k\}$, such that labels with similar $z$ are assigned to the same discrete label $c$, see \figref{fig:split}. The discrete labels may be binary ($k=2$) or multiclass ($k>2$). This allows us to use standard information gain criteria based on Shannon entropy or Gini impurity as defined in \eqnref{eqn:infogain}. Critically, discretization is performed independently when training each node and depends on the distribution of labels at a given node (contrast with \cite{Lim2013CVPR}).

We consider two straightforward approaches to obtaining the discrete label set $\C$ given $\Z$. Our first approach is to cluster $z$ into $k$ clusters using K-means (projecting $z$ onto 5 dimensions prior to clustering). Alternatively, we can quantize $z$ based on the top $\log_2(k)$ PCA dimensions, assigning $z$ a discrete label $c$ according to the orthant (generalization of quadrant) into which $z$ falls. Both approaches perform similarly but the latter is slightly faster. We use PCA quantization to obtain $k=2$ labels unless otherwise specified.

\subsection{Ensemble Model}\label{sec:structured:ensemble}

Finally, we define how to combine a set of $n$ labels $y_1 \ldots y_n$ into a single prediction for both training (to set leaf labels) and testing (to merge predictions). As before, we sample an $m$ dimensional mapping $\map_\phi$ and compute $z_i = \map_\phi(y_i)$ for each $i$. We select the label $y_k$ whose $z_k$ is the medoid, \ie the $z_k$ that minimizes the sum of distances to all other $z_i$\footnote{The medoid $z_k$ minimizes $\sum_{ij} (z_{kj} - z_{ij})^2$. This is equivalent to $\min_k\sum_{j} (z_{kj} - \bar{z}_j)^2$ and can be computed efficiently in time $O(nm)$.}. Note that typically we only need to compute the medoid for small $n$ (either for training a leaf node or merging the output of multiple trees), hence using a coarse distance metric suffices. 

\hide{\eqns{\normalsize}{
 & & \min_k \sum_{ij} (z_{kj} - z_{ij})^2 \\
 &=& \min_k \sum_{ij} (z_{kj}^2  -2z_{kj}z_{ij} + z_{ij}^2) \\
 &=& \min_k \sum_{ij} (z_{kj}^2  -2z_{kj}z_{ij}) \\
 &=& \min_k \sum_{j} (z_{kj}^2  -2z_{kj}\bar{z}_j)n \\
 &=& \min_k \sum_{j} (z_{kj}^2  -2z_{kj}\bar{z}_j + \bar{z}_j^2) \\
 &=& \min_k \sum_{j} (z_{kj} - \bar{z}_j)^2
}}

The biggest limitation is that any prediction $y\in\Y$ must have been observed during training; the ensemble model is unable to synthesize novel labels. Indeed, this is impossible without additional information about $\Y$. In practice, domain specific ensemble models are preferable. For example, in edge detection we apply structured prediction to obtain edge maps for each image patch independently and merge overlapping predictions by averaging (note that in this case structured prediction operates at the patch level and not the image level). 

\section{Edge Detection}\label{sec:edges}

We now describe how to apply our structured forests to edge detection. As input our method takes an image that may contain multiple channels, such as an RGB or RGBD image. The task is to label each pixel with a binary variable indicating whether the pixel contains an edge or not. Similar to the task of semantic image labeling \cite{Kontschieder2011ICCV}, the labels within a small image patch are highly interdependent, providing a promising candidate problem for our structured forest approach.

We assume we are given a set of segmented training images, in which the boundaries between the segments correspond to contours \cite{Arbelaez2011PAMI,Silberman2012ECCV}. Given an image patch, its annotation can be specified either as a \textit{segmentation mask} indicating segment membership for each pixel (defined up to a permutation) or a binary \textit{edge map}. We use $y\in\Y=\mathbb{Z}^{d\times d}$ to denote the former and $y'\in\Y'=\{0,1\}^{d\times d}$ for the latter, where $d$ indicates patch width. An edge map $y'$ can always be trivially derived from segmentation mask $y$, but not vice versa. We utilize both representations in our approach.

Next, we describe how we compute the input features $x$, the mapping functions $\map_\phi$ used to determine splits, and the ensemble model used to combine multiple predictions.

\vspace{5pt}\noindent\textbf{Input features:} Our learning approach predicts a structured $16 \times 16$ segmentation mask from a larger $32 \times 32$ image patch. We begin by augmenting each image patch with multiple additional \textit{channels} of information, resulting in a feature vector $x\in\R^{32\times32\times K}$ where $K$ is the number of channels. We use features of two types: pixel lookups $x(i,j,k)$ and pairwise differences $x(i_1,j_1,k)-x(i_2,j_2,k)$, see \secref{sec:forests}.

Inspired by Lim \etal \cite{Lim2013CVPR}, we use a similar set of color and gradient channels (originally developed for fast pedestrian detection \cite{Dollar2010BMVC}). We compute 3 color channels in CIE-LUV color space along with normalized gradient magnitude at 2 scales (original and half resolution). Additionally, we split each gradient magnitude channel into 4 channels based on orientation. The result is 3 color, 2 magnitude and 8 orientation channels, for a total of $13$ channels.

We blur the channels with a radius 2 triangle filter and downsample by a factor of 2, resulting in $32\cdot32\cdot13/4 = 3328$ candidate  features $x$. Motivated by \cite{Lim2013CVPR}, we also compute pairwise difference features. We apply a large triangle blur to each channel (8 pixel radius), and downsample to a resolution of $5 \times 5$. Sampling all candidate pairs and computing their differences yields an additional $\binom{5\cdot5}{2} = 300$ candidate features per channel, resulting in 7228 total candidate features.

\vspace{5pt}\noindent\textbf{Mapping function:} To train decision trees, we need to define a mapping $\map:\Y\rightarrow\Z$ as described in \secref{sec:structured}. Recall that our structured labels $y$ are $16 \times 16$ segmentation masks. One option is to use $\map:\Y\rightarrow\Y'$, where $y'$ represents the binary edge map corresponding to $y$. Unfortunately Euclidean distance over $\Y'$ yields a brittle distance measure.

We therefore define an alternate mapping $\map$. Let $y(j)$ for $1\le j\le256$ denote the segment index of the $j^{th}$ pixel of $y$. Individually a single value $y(j)$ yields no information about $y$, since $y$ is defined only up to a permutation. Instead we can sample a pair of locations $j_1\ne j_2$ and check if they belong to the same segment, $y(j_1)=y(j_2)$. This allows us to define $z=\map(y)$ as a large binary vector that encodes $[y(j_1)=y(j_2)]$ for every unique pair of indices $j_1\ne j_2$. While $\Z$ has $\binom{256}{2}$ dimensions, in practice we only compute a subset of $m$ dimensions as discussed in \secref{sec:structured:information}. We found a setting of $m=256$ and $k=2$ gives good results, effectively capturing the similarity of segmentation masks.

\vspace{5pt}\noindent\textbf{Ensemble model:} Random forests achieve robust results by combining the output of multiple trees. While merging segmentation masks $y\in\Y$ for overlapping patches is difficult, multiple overlapping edge maps $y'\in\Y'$ can be averaged to yield a soft edge response. Thus in addition to the learned mask $y$, we also store the corresponding edge map $y'$ at each leaf node, thus allowing predictions to be combined quickly and simply through averaging during inference.

\vspace{5pt}\noindent\textbf{Efficiency:} The surprising efficiency of our approach derives from the use of structured labels that predict information for an entire image neighborhood. This greatly reduces the number of trees $T$ that need to be evaluated. We compute our structured output densely on the image with a stride of 2 pixels, thus with $16 \times 16$ output patches, each pixel receives $16^2T/4 \approx 64T$ predictions. In practice we use $T=4$ and thus the score of each pixel in the output edge map is averaged over 256 votes.

A critical assumption is that predictions are uncorrelated. Since both the inputs and outputs of each tree overlap, we train $2T$ total trees and evaluate an alternating set of $T$ trees at each adjacent location. Use of such a `checkerboard pattern' improves results somewhat, introducing larger separation between the trees did not improve results further.

\subsection{Multiscale Detection (SE+MS)}\label{sec:edges:ms}

We now describe the first of two enhancements to our base algorithm. Inspired by the work of Ren \cite{Ren2008ICCV}, we implement a multiscale version of our edge detector. Given an input image $I$, we run our structured edge detector on the original, half, and double resolution version of $I$ and average the result of the three edge maps after resizing to the original image dimensions. Although somewhat inefficient, the approach noticeably improves edge quality. We refer to the multiscale version of our structured edge detector as SE+MS.

\subsection{Edge Sharpening (SE+SH)}\label{sec:edges:sh}

We observed that predicted edge maps from our structured edge detector are somewhat diffuse. For strong, isolated edges non-maximal suppression can be used to effectively detect edge peaks. However, given fine image structures edge responses can `bleed' together resulting in missed detections; likewise, for weak edges that receive few votes no clear peak may emerge. The underlying cause for the diffuse edge responses is that the individually predicted edge maps are noisy and are not perfectly aligned to each other or the underlying image data. Specifically, each overlapping prediction may be shifted by a few pixels from the true edge location.

To address this phenomenon, we introduce a new \textit{sharpening} procedure that aligns edge responses from overlapping predictions. Our core observation is that local image color and depth values can be used to more precisely localize predicted responses. Intuitively, given a predicted segmentation mask, the mask can be morphed slightly so that it better matches the underlying image patch. Aligning overlapping masks to the underlying image data implicitly aligns the masks with each other, resulting in sharper, better localized edge responses.

Sharpening takes a predicted segmentation mask $y \in \Y$ and the corresponding image patch $x \in \X$ and produces a new mask that better aligns to $x$. As before, let $y(j)$ denote the segment index of the $j^{th}$ pixel of mask $y$. First, for each segment $s$, we compute its mean color $\mu_s=E[x(j)|y(j)=s]$ using all pixels $j$ in $s$. Next, we iteratively update the assigned segment for each pixel by assigning it to the segment which minimizes $\|\mu_s-x(j)\|_2$. For each pixel we restrict the set of assignable segments to the segments that are immediately adjacent to it (4-connected neighborhood). Given the new sharpened segmentation masks, we compute and average their corresponding edge maps as before. However, since the edge maps are better aligned to the image data the resulting aggregated edge map is sharper.

Sharpening can be repeated multiple times prior to averaging the corresponding edge maps. Experiments reveal that the first sharpening step produces the largest gains, and in practice two steps suffice. Taking advantage of the sparsity of edges, the sharpening procedure can be implemented efficiently. For details we direct readers to source code. Note that sharpening is not guaranteed to improve results but we find it is quite effective in practice. We refer to the sharpened versions of our structured edge detector as SE+SH and SE+MS+SH.

\begin{figure*}\center
\begin{tabular}{r@{\hskip 1mm}c@{\hskip 1mm}c@{\hskip 1mm}c@{\hskip 1mm}c@{\hskip 1mm}c}
  \bsdsFigHelper{}{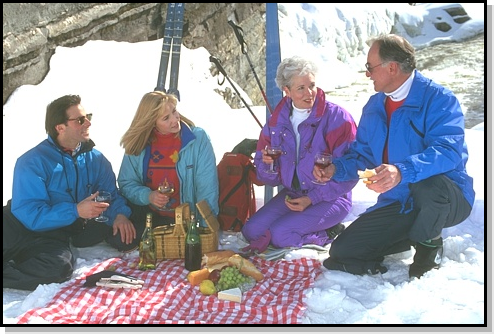}
  \bsdsFigHelper{ground truth}{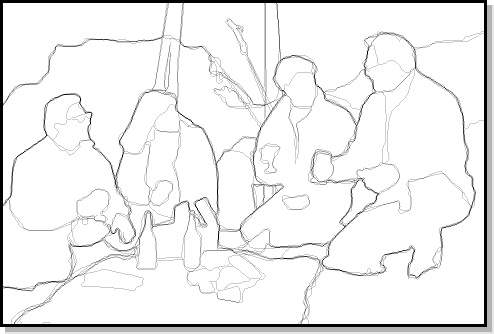}
  \bsdsFigHelper{gPb+owt+ucm}{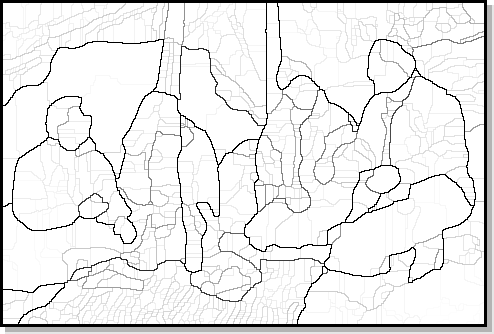}
  \bsdsFigHelper{SketchTokens}{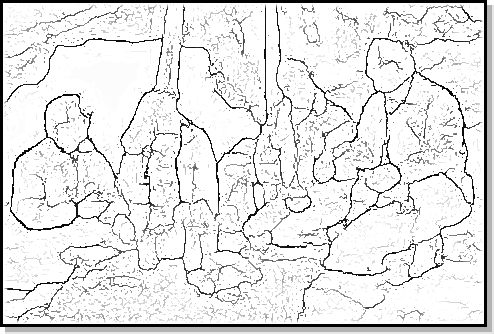}
  \bsdsFigHelper{SCG}{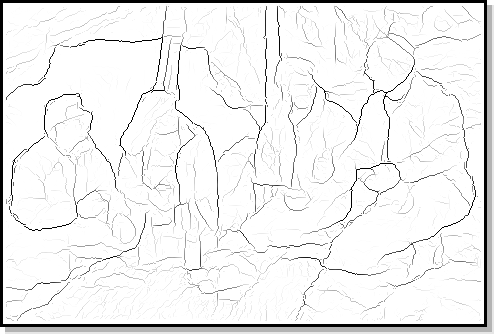}
  \bsdsFigHelper{SE}{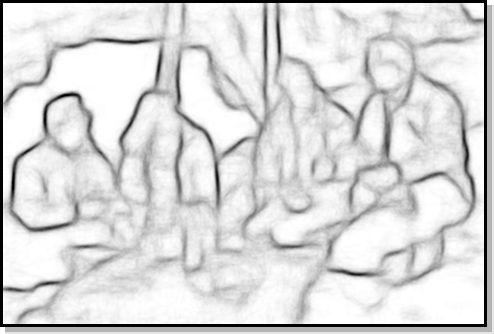}
  \bsdsFigHelper{SE+MS}{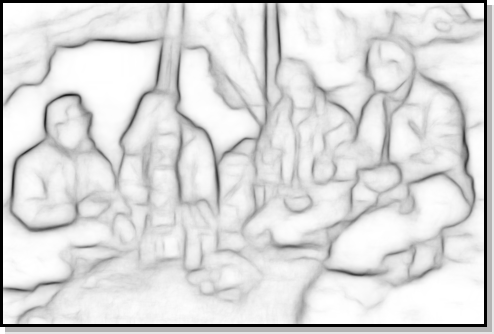}
  \bsdsFigHelper{SE+SH}{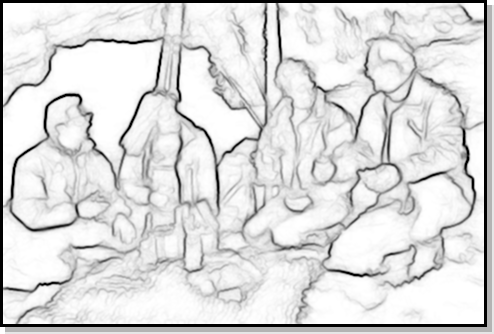}
  \bsdsFigHelper{SE+MS+SH}{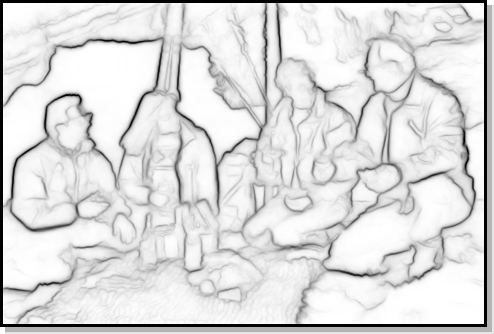}
\end{tabular}
\Caption{Illustration of edge detection results on the BSDS500 dataset on five sample images. The first two rows show the original image and ground truth. The next three rows contain results for gPb-owt-ucm \cite{Arbelaez2011PAMI}, Sketch Tokens \cite{Lim2013CVPR}, and SCG \cite{Ren2012NIPS}. The final four rows show our results for variants of SE. Use viewer zoom functionality to see fine details. }\label{fig:res:bsds}
\end{figure*}

\begin{figure*}\center
\begin{tabular}{r@{\hskip 1mm}c@{\hskip 1mm}c@{\hskip 1mm}c@{\hskip 1mm}c@{\hskip 1mm}c}
  \bsdsFigHelper{}{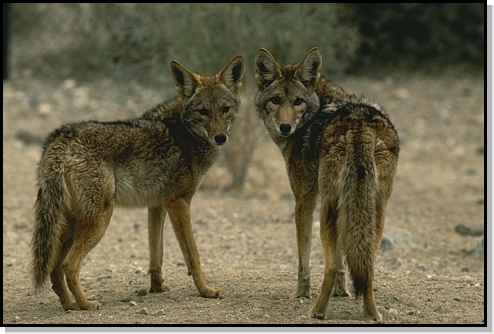}
  \bsdsFigHelper{ground truth}{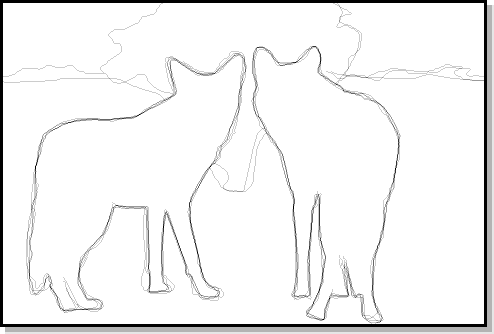}
  \bsdsFigHelper{high precision}{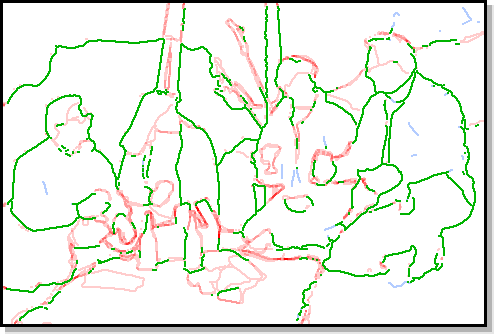}
  \bsdsFigHelper{optimal thresh}{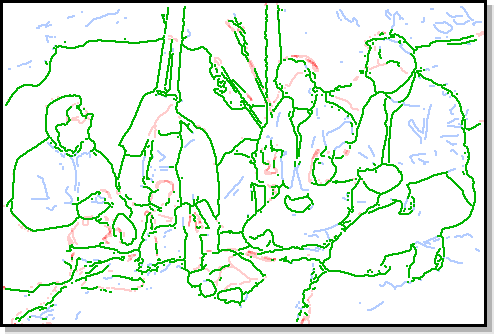}
  \bsdsFigHelper{high recall}{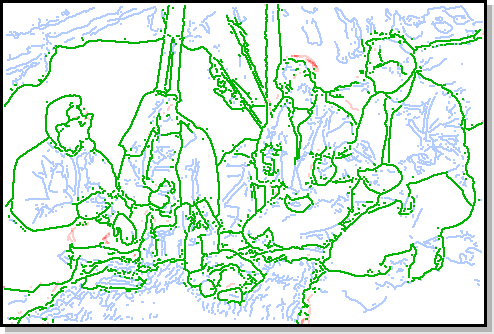}
\end{tabular}\vspace{-1mm}
\Caption{Visualizations of matches and errors of SE+MS+SH compared to BSDS ground truth edges. Edges are thickened to two pixels for better visibility; the color coding is \textcolor[rgb]{0,.7,0}{green}=true positive, \textcolor[rgb]{.4,.5,1}{blue}=false positive, \textcolor[rgb]{1,.3,.3}{red}=false negative. Results are shown at three thresholds: high precision (T$\approx$.26, P$\approx$0.88, R=.50), ODS threshold (T$\approx$.14, P=R$\approx$.75), and high recall (T$\approx$.05, P=.50, R$\approx$0.93).}\label{fig:res:bsdseval}\vspace{-2mm}
\end{figure*}

\section{Results}\label{sec:results}

In this section we analyze the performance of our structured edge (SE) detector in detail. First we analyze the influence of parameters in \secref{sec:results:params} and test SE variants in \secref{sec:results:variants}. Next, we compare results on the BSDS \cite{Arbelaez2011PAMI} and NYUD \cite{Silberman2012ECCV} datasets to the state-of-the-art in \secref{sec:results:bsds} and \secref{sec:results:nyu}, respectively, reporting both accuracy and runtime. We conclude by demonstrating the cross dataset generalization of our approach in \secref{sec:results:cross}.

The majority of our experiments are performed on the Berkeley Segmentation Dataset and Benchmark (BSDS500) \cite{Martin2001ICCV,Arbelaez2011PAMI}. The dataset contains 200 training, 100 validation, and 200 testing images. Each image has hand labeled ground truth contours. Edge detection accuracy is evaluated using three standard measures: fixed contour threshold (ODS), per-image best threshold (OIS), and average precision (AP) \cite{Arbelaez2011PAMI}. To evaluate accuracy in the high recall regime, we additionally introduce a new measure, recall at 50\% precision (R50), in \secref{sec:results:variants}. Prior to evaluation, we apply a standard non-maximal suppression technique to our edge maps to obtain thinned edges \cite{Canny1986PAMI}. Example detections on BSDS are shown in \figref{fig:res:bsds} and visualizations of edge accuracy are shown in \figref{fig:res:bsdseval}.

\subsection{Parameter Sweeps}\label{sec:results:params}

We set all parameters with the help of the BSDS validation set which is fully independent of the test set. Parameters include: structured forest splitting parameters (\eg, $m$ and $k$), feature parameters (\eg, image and channel blurring), and model and tree parameters (\eg number of trees and data quantity). Training takes \app20 minute per tree using one million patches and is parallelized over trees. Evaluation of trees is parallelized as well, we use a quad-core machine for all reported runtimes.

\begin{figure}\center
\begin{tabular}{c@{\hskip .05in}c}
  \incp{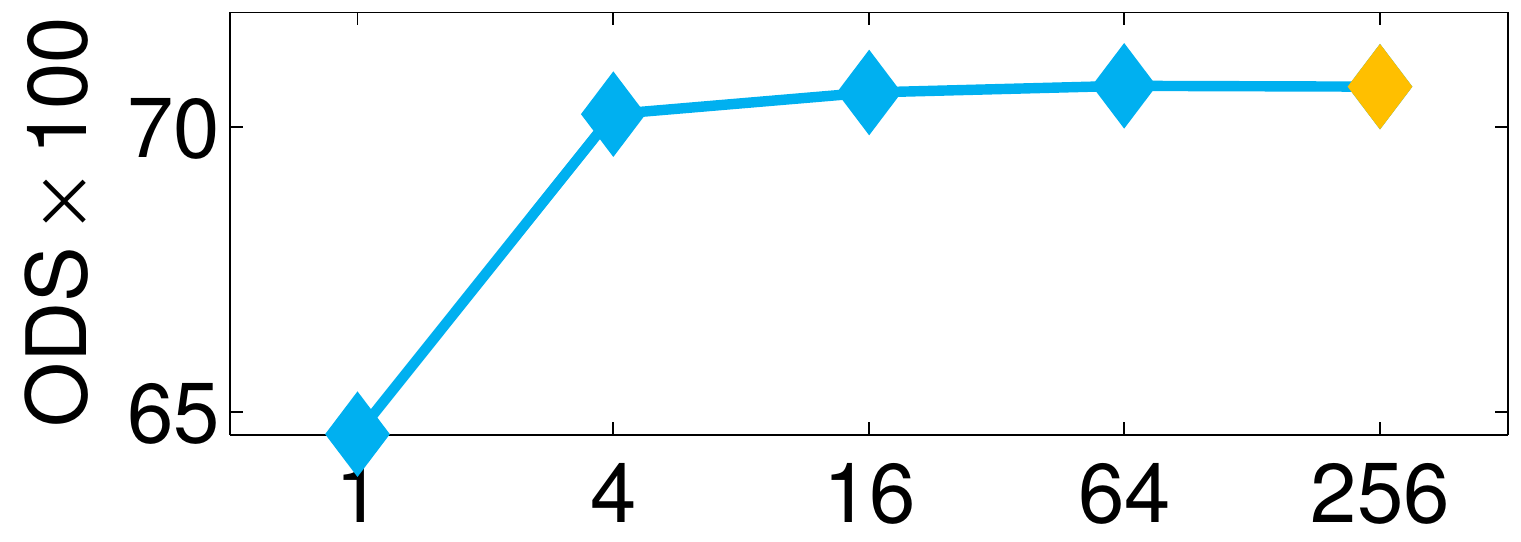} & \incp{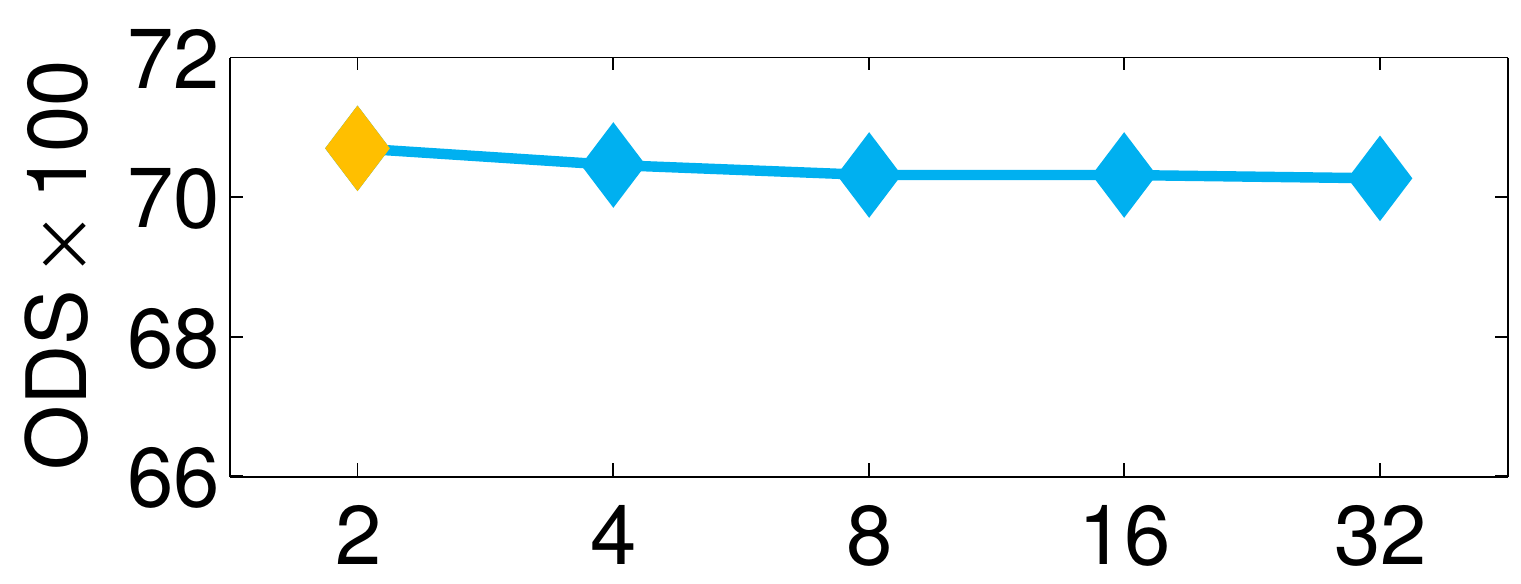} \\
  (a) $m$ (size of $\Z$) & (b) $k$ (size of $\C$) \\
  \incp{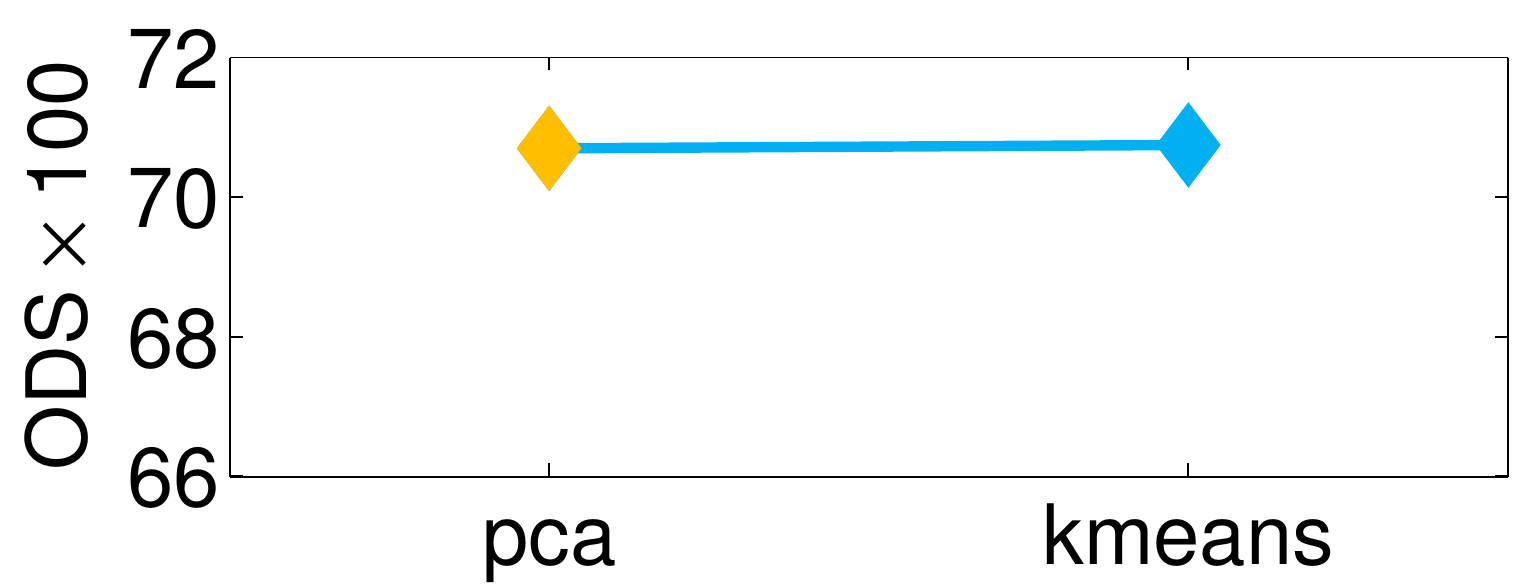} & \incp{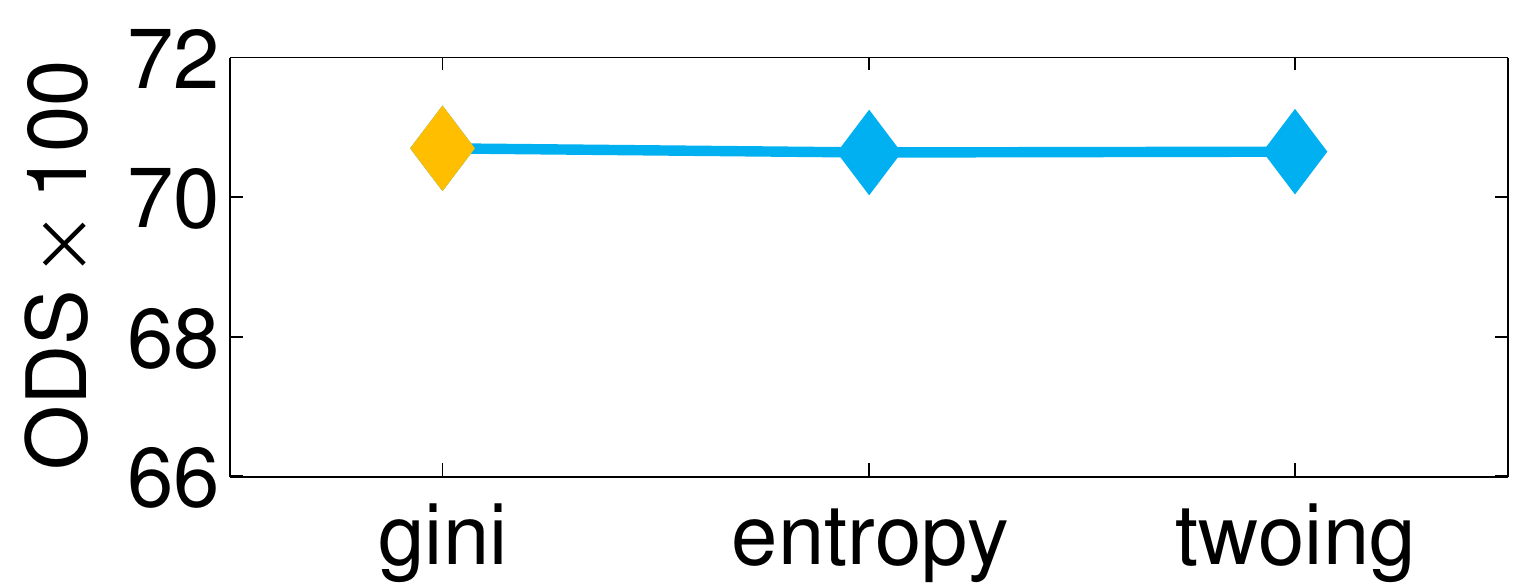} \\
  (c) discretization type & (d) information gain \\
  \end{tabular}\vspace{1mm}
  \Caption{ Splitting parameter sweeps. See text for details. }
  \label{fig:sweeps:split}\vspace{-1mm}
\end{figure}

\begin{figure}\center
\begin{tabular}{c@{\hskip .05in}c}
  \incp{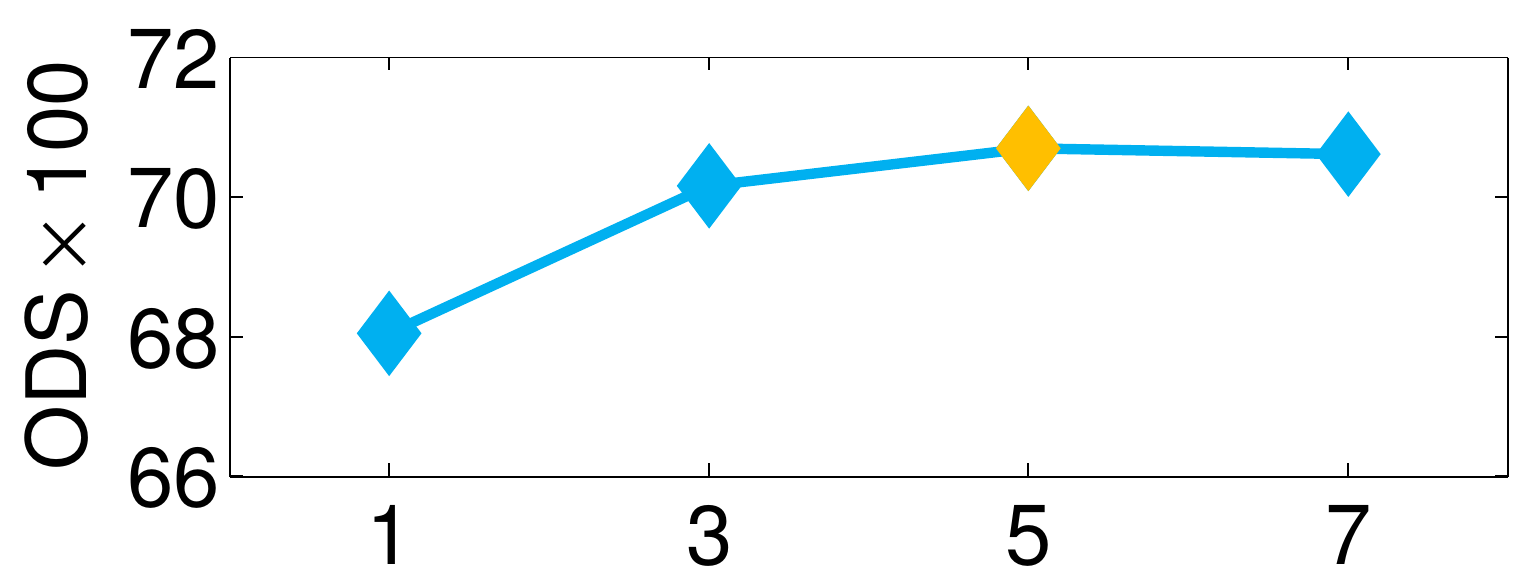} & \incp{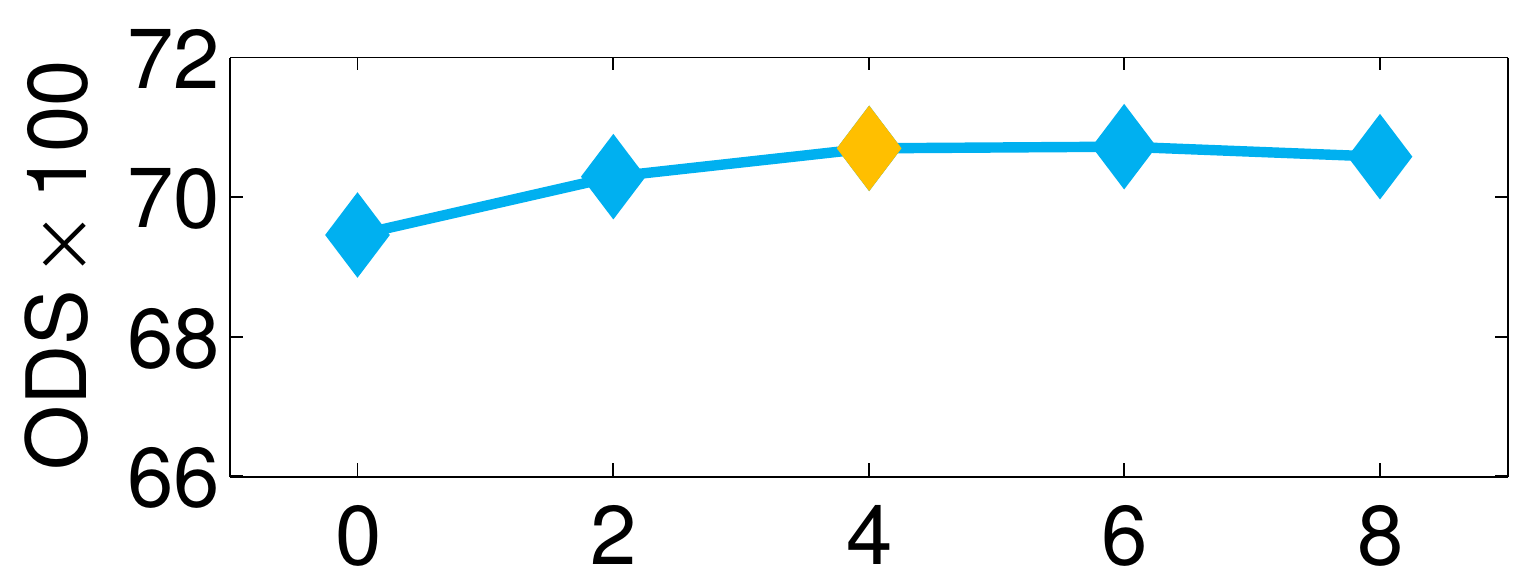} \\
  (a) \# grid cells & (b) \# gradient orients \\
  \incp{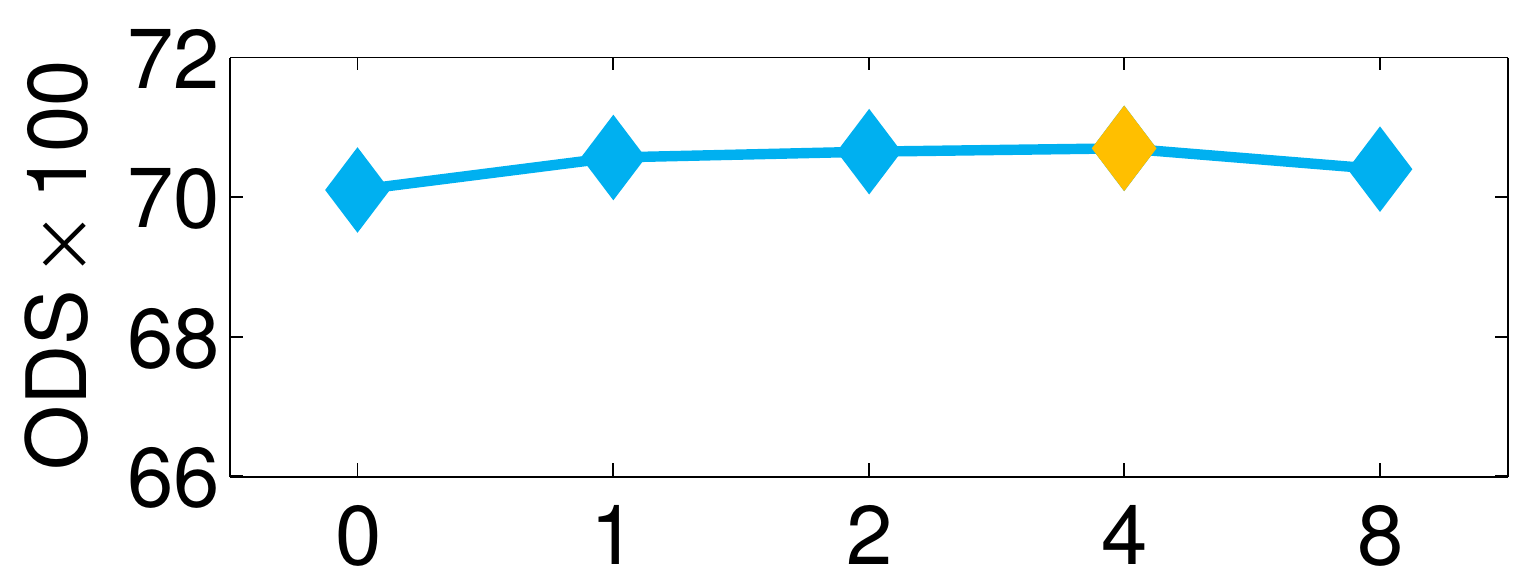} & \incp{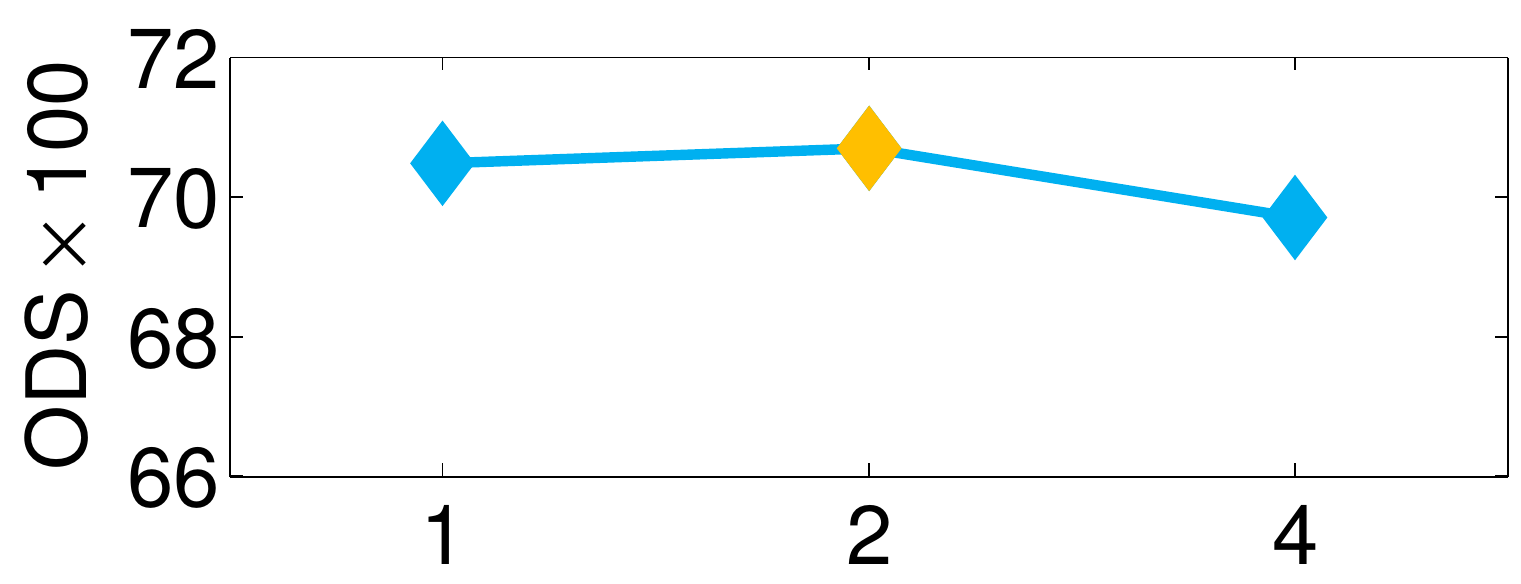} \\
  (c) normalization radius & (d) channel downsample \\
  \incp{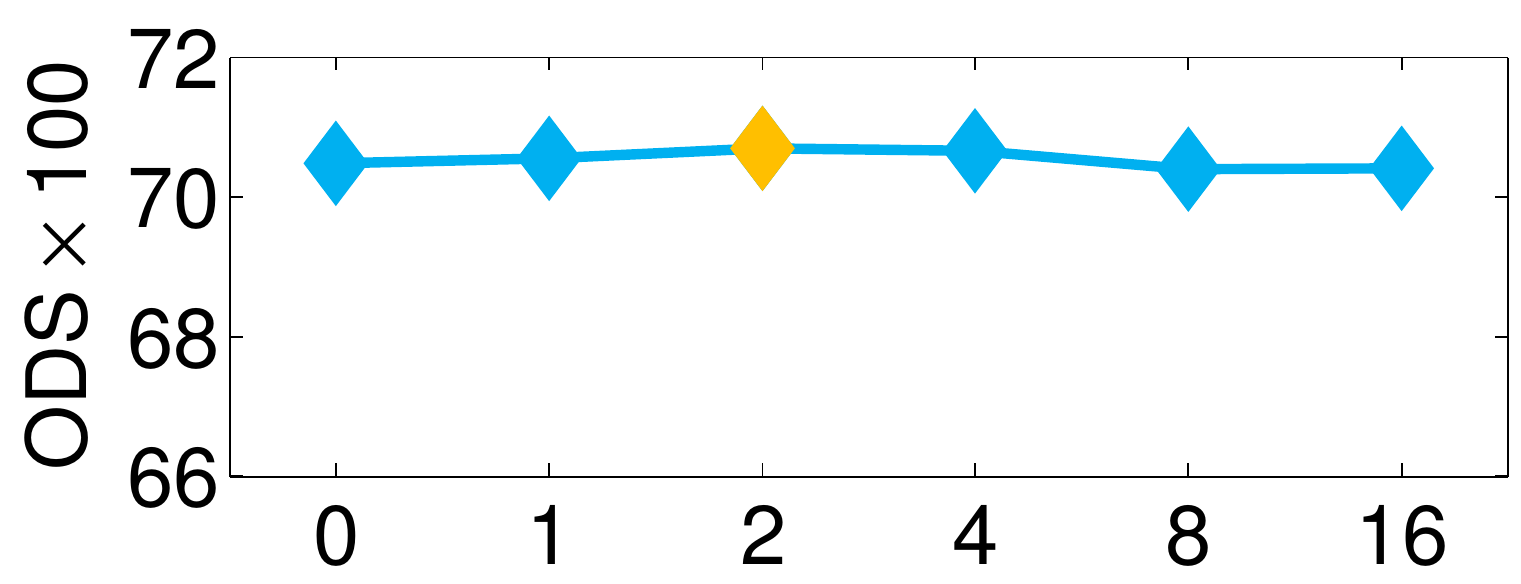} & \incp{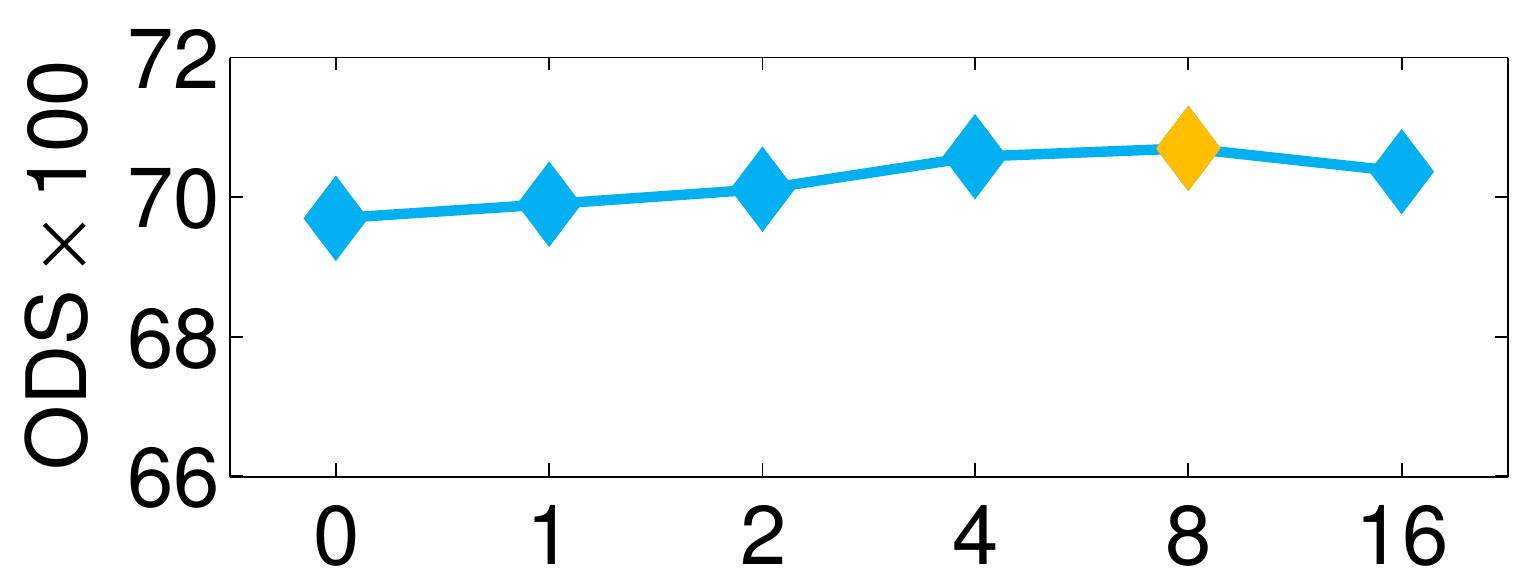} \\
  (e) channel blur & (f) self-similarity blur
  \end{tabular}\vspace{1mm}
  \Caption{ Feature parameter sweeps. See text for details. }
  \label{fig:sweeps:features}
\end{figure}

\begin{figure}\center
\begin{tabular}{c@{\hskip .05in}c}
  \incp{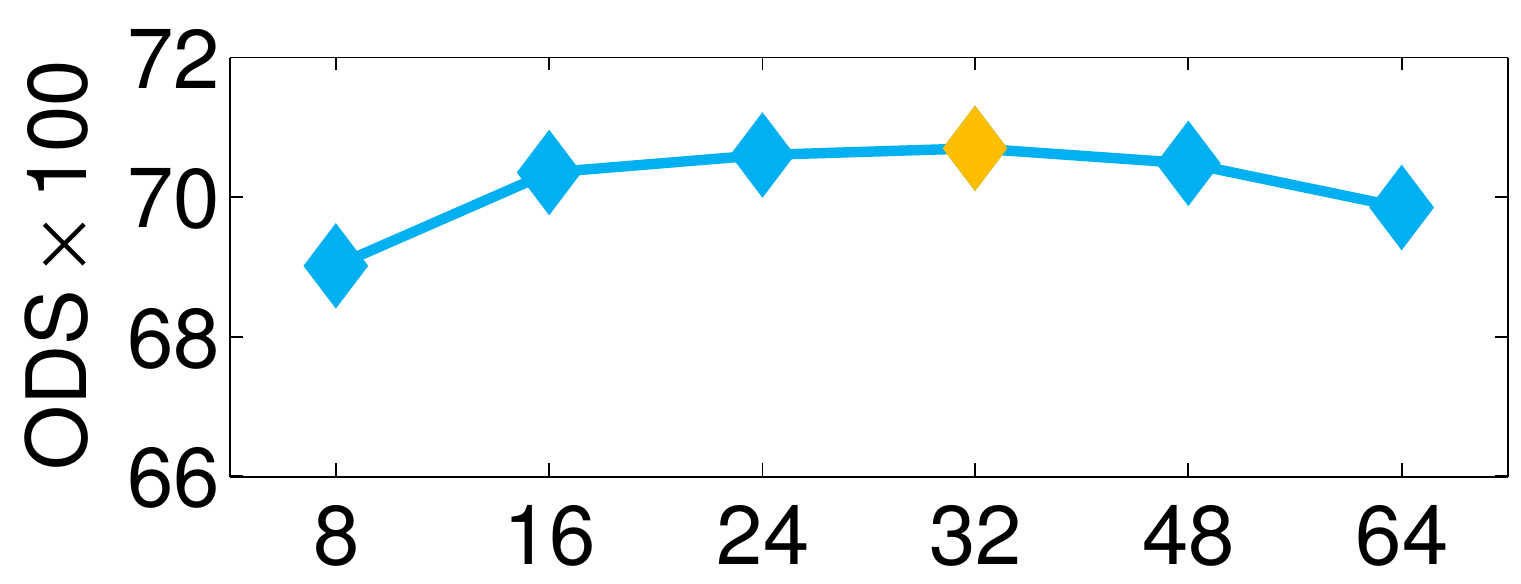} & \incp{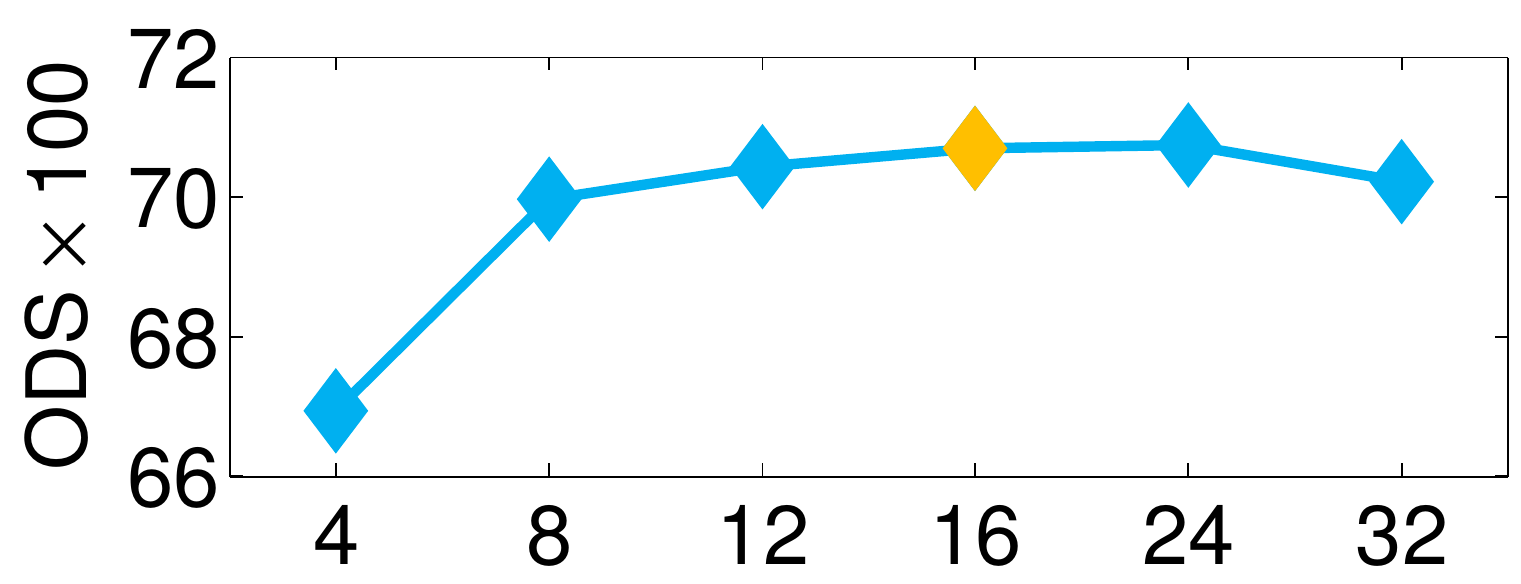} \\
  (a) patch size for $x$ & (b) patch size for $y$ \\
  \incp{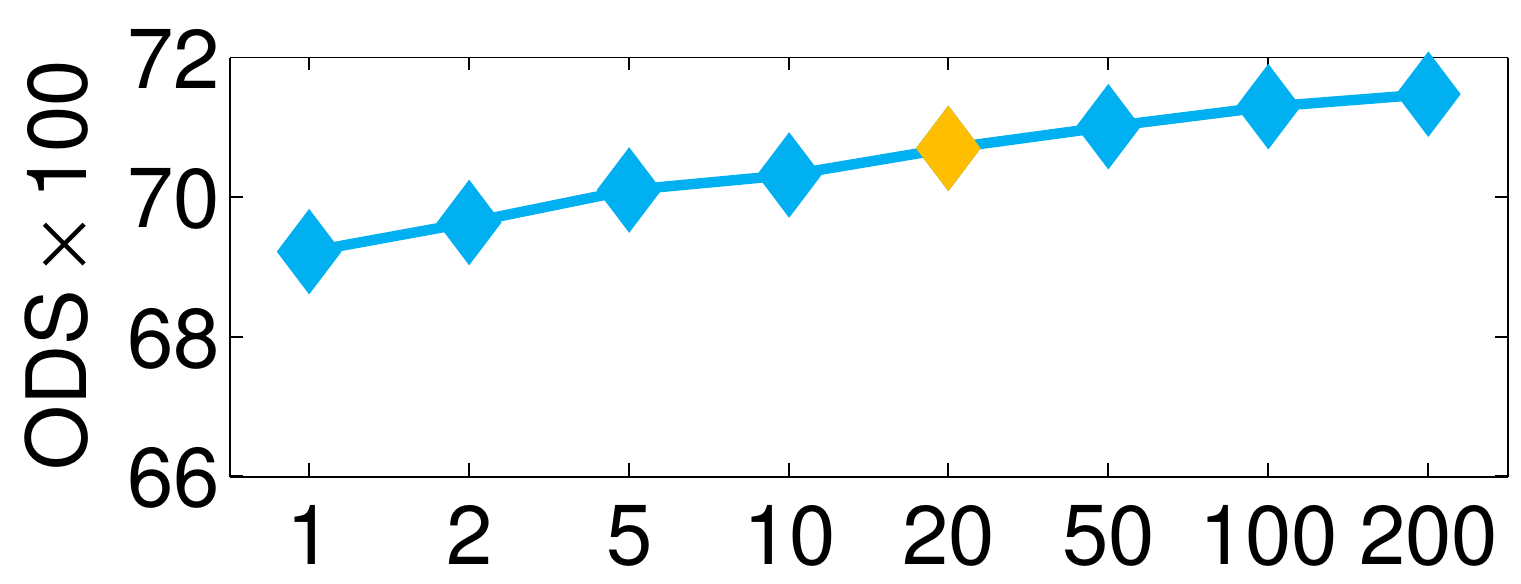} & \incp{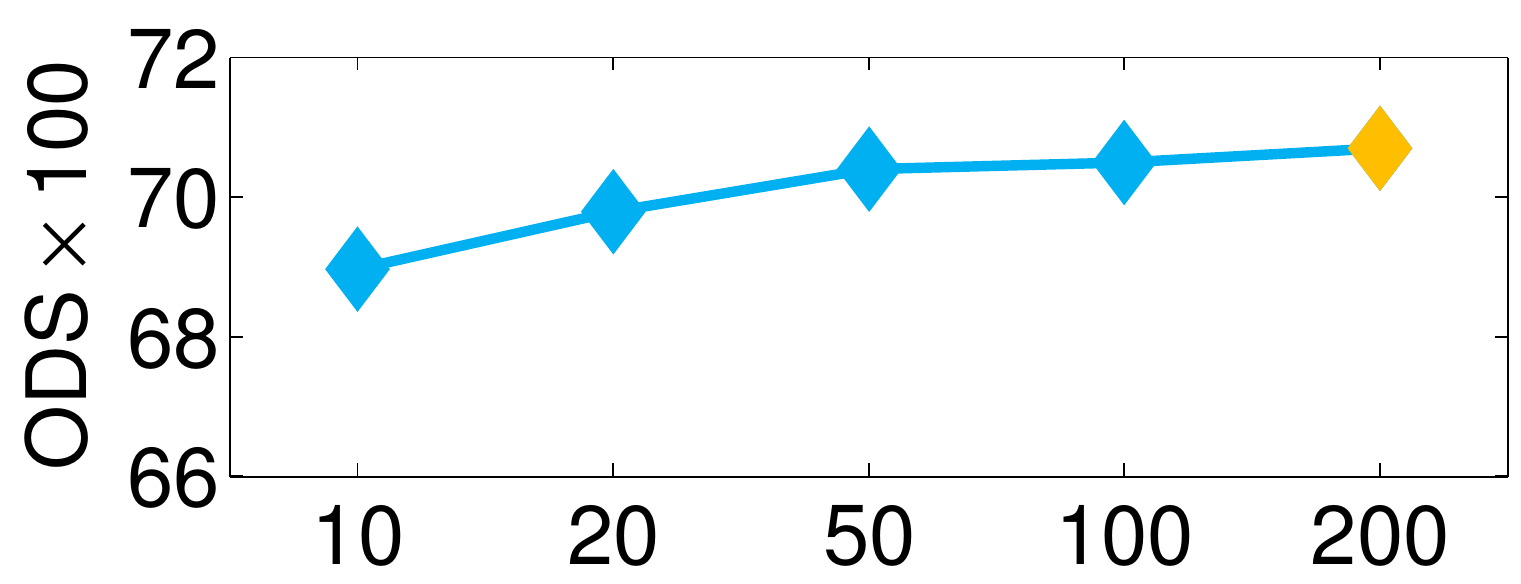} \\
  (c) \# train patches $\times 10^4$ & (d) \# train images \\
  \incp{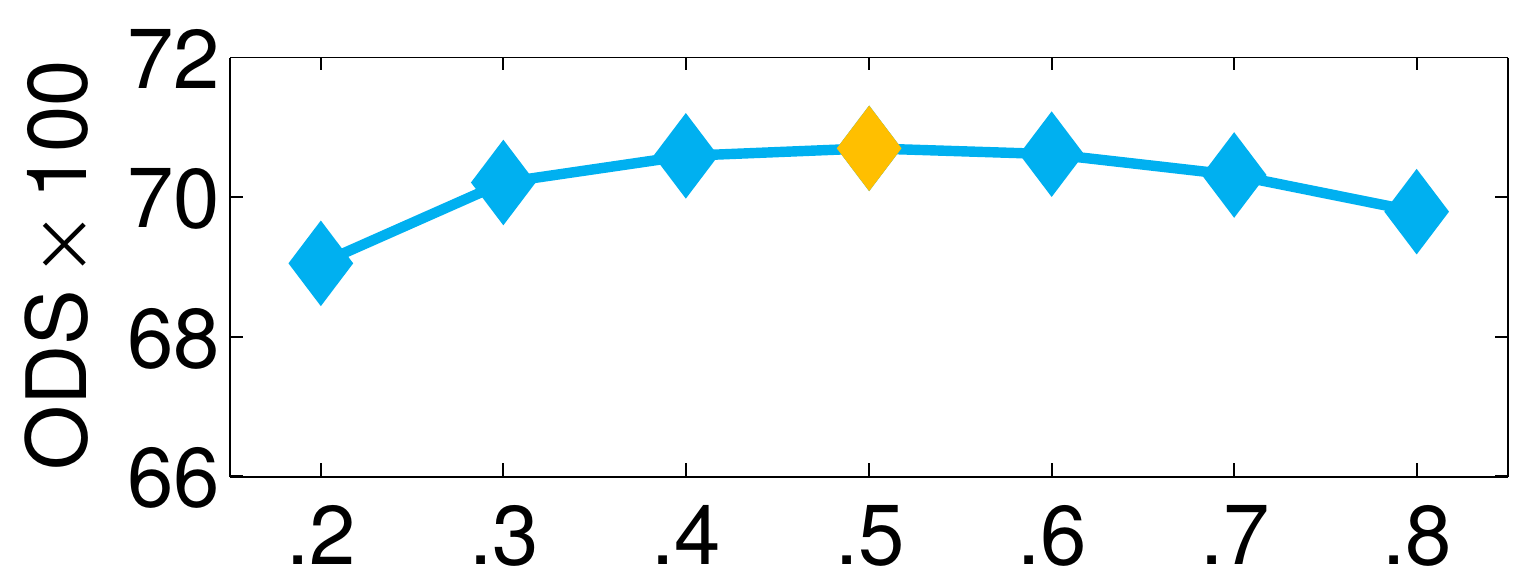} & \incp{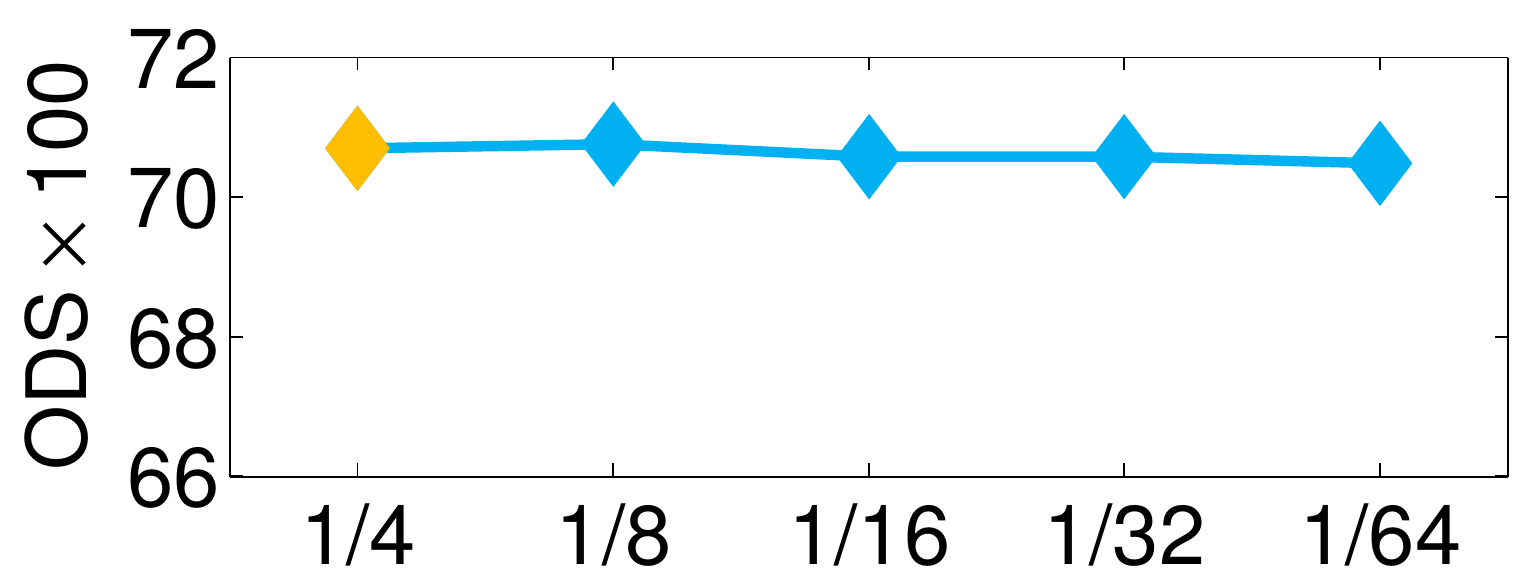} \\
  (e) fraction `positives' & (f) fraction features \\
  \incp{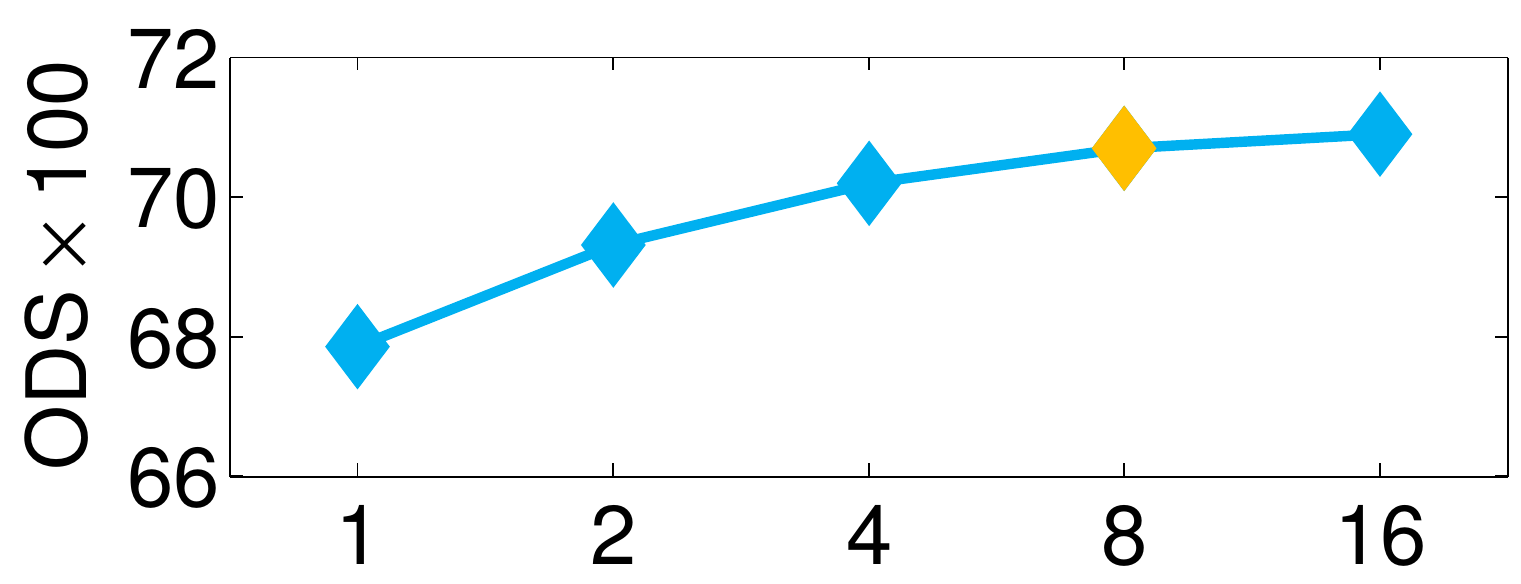} & \incp{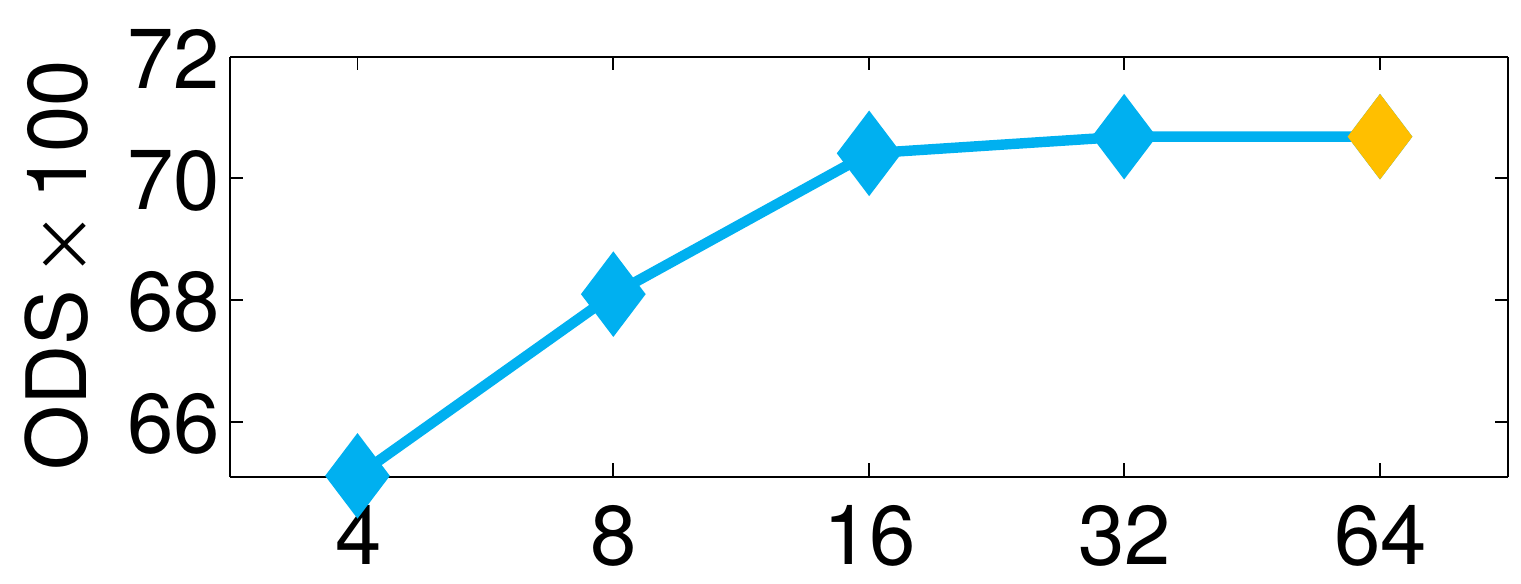} \\
  (g) \# decision trees & (h) max tree depth \\
  \incp{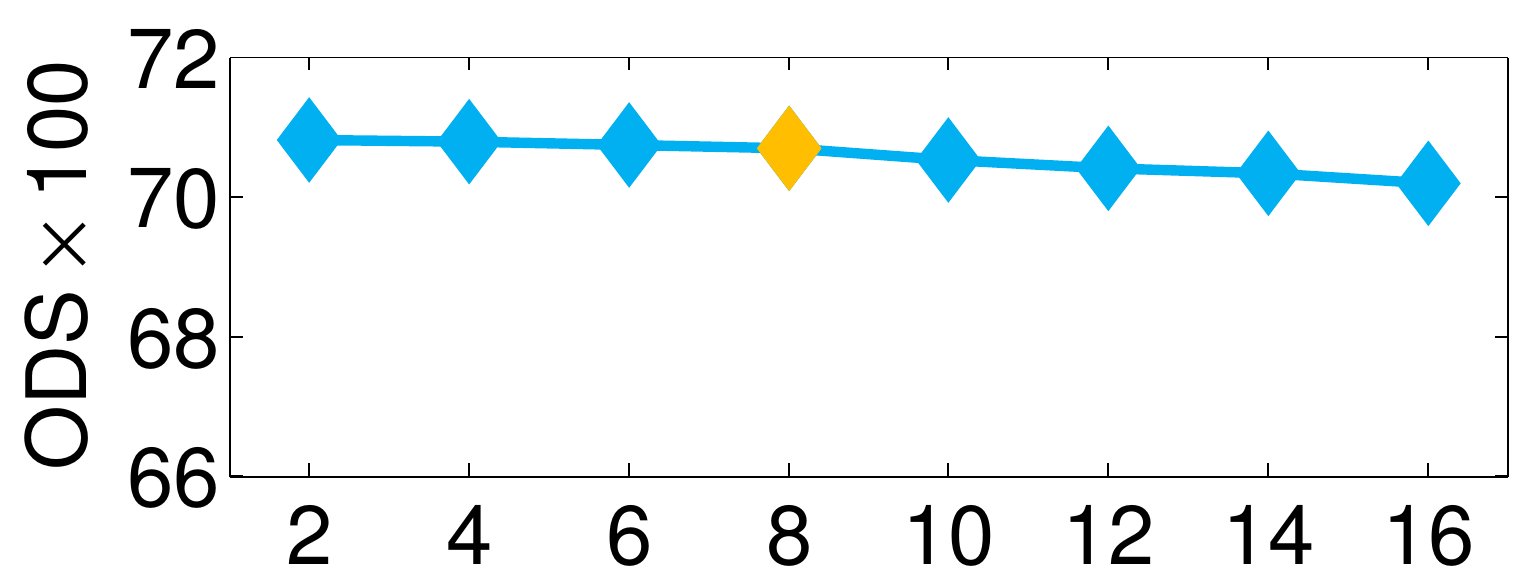} & \incp{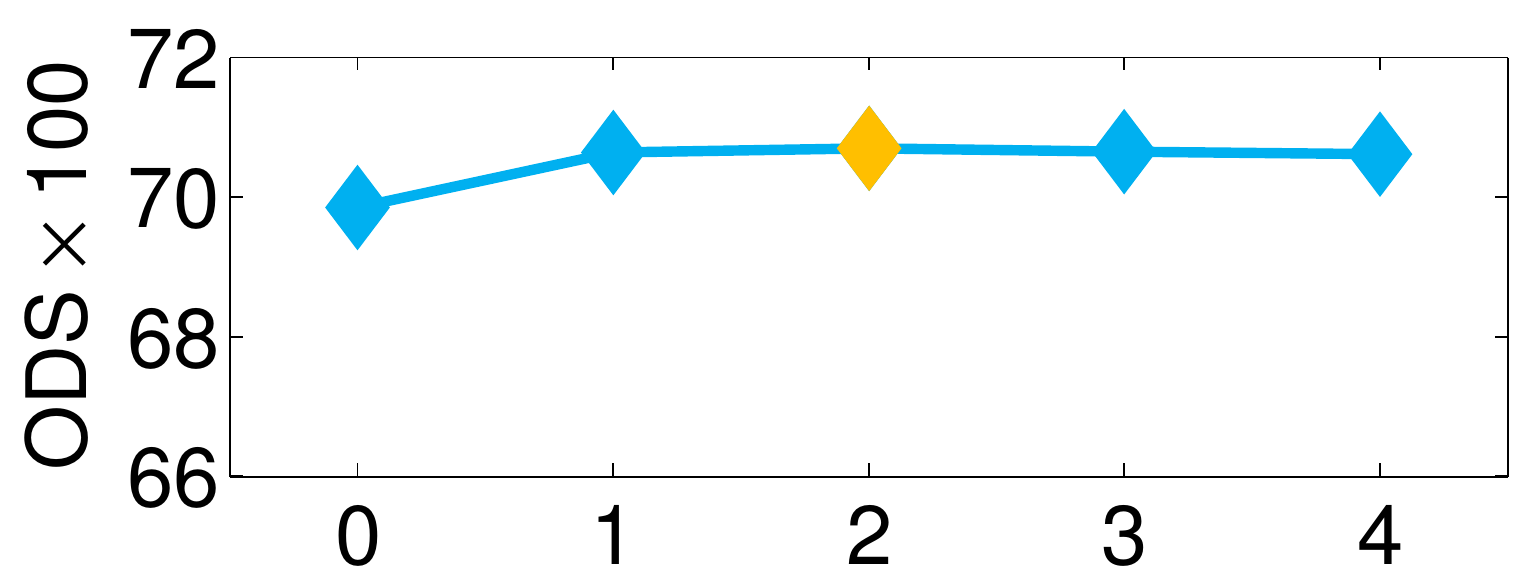} \\
  (i) min samples per node & (j) \# sharpening steps
  \end{tabular}\vspace{1mm}
  \Caption{ Model parameter sweeps. See text for details. }
  \label{fig:sweeps:model}\vspace{5mm}
\end{figure}

In \figsref{fig:sweeps:split}-\ref{fig:sweeps:model} we explore the effect of choices of \textbf{splitting}, \textbf{model} and \textbf{feature} parameters. For each experiment we train on the 200 image training set and measure edge detection accuracy on the 100 image validation set (using the standard ODS performance metric). All results are averaged over 5 trials. First, we set all parameters to their default values indicated by orange markers in the plots. Then, keeping all but one parameter fixed, we explore the effect on edge detection accuracy as a single parameter is varied.

Since we explore a large number of parameters settings, we perform our experiments using a slightly reduced accuracy model that is faster to train. Specifically we train using fewer patches ($2\cdot10^5$ versus $10^6$) and utilize sharpening (SH) but not multiscale detection (MS). Also, the validation set is more challenging than the test set and we evaluate using 25 thresholds instead of 99, further reducing accuracy (.71 ODS). Finally, we note that sweep details have changed slightly from the our previous work \cite{Dollar2013ICCV}; most notably, the sweeps now utilize sharpening but not multiscale detection.

\textbf{Splitting Parameters:} In \figref{fig:sweeps:split} we explore how best to measure information gain over structured labels. Recall we utilize a two-stage approach of mapping $\Y\rightarrow\Z$ followed by $\Z\rightarrow\C$. Plots (a) and (b) demonstrate that $m=|\Z|$ should be large and $k=|\C|$ small. Results are robust to both the discretization method and the discrete measure of information gain as shown in plots (c) and (d).

\textbf{Feature Parameters:} \figref{fig:sweeps:features} shows how varying the channel features affects accuracy. We refer readers to \secref{sec:edges} and source code for details, here we only note that performance is relatively insensitive to a broad range of parameter settings.

\textbf{Model Parameters:} In \figref{fig:sweeps:model} we plot the influence of parameters governing the model and training data. (a) and (b) show the effect of image and label patch sizes on accuracy, $32\times32$ image patches and $16\times16$ label patches are best. (c) and (d) show that increasing the number of patches and training images improves accuracy. (e) shows that about half the sampled patches should be `positive' (have more than one ground truth segment) and (f) shows that training each tree with a fraction of total features has negligible impact on accuracy (but results in proportionally lower memory usage). In (g)-(i) we see that many, deep, un-pruned trees give best performance (nevertheless, we prune trees so every node has at least 8 training samples to decrease model size). Finally (j) shows that two sharpening steps give best results. Impact of sharpening is explored in more detail next in \secref{sec:results:variants}.

\subsection{Structured Edge Variants}\label{sec:results:variants}

Given our trained detector, sharpening (SH) and multiscale detection (MS) can be used to enhance results. In this section we analyze the performance of the four resulting combinations. We emphasize that the same trained model can be used with sharpening and multiscale detection enabled at runtime.

\begin{figure} \center
  \includegraphics[width=.42\textwidth]{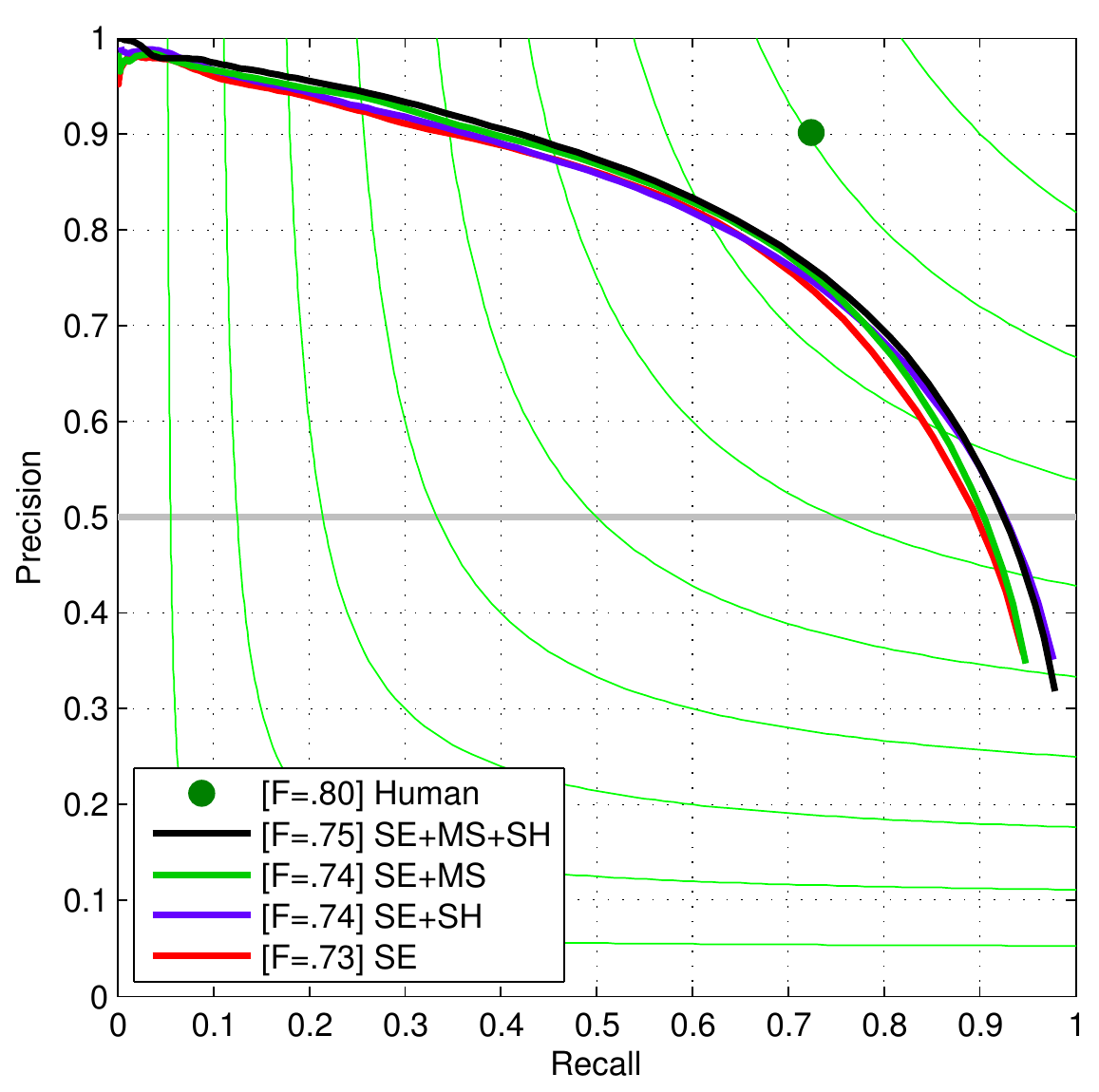}
\Caption{Results of structured edges (SE) with sharpening (+SH) and multiscale detection (+MS). SH increases recall while MS increases precision; their combination gives best results.}
 \label{fig:pr:bsds-variants}
\end{figure}

In \figref{fig:pr:bsds-variants} we plot precision/recall curves for the four variants of our approach: SE, SE+MS, SE+SH, and SE+MS+SH. Summary statistics are reported in the bottom rows of \tableref{table:res:bsds}. SE has an ODS score of .73 and SE+MS and SE+SH both achieve an ODS of .74. SE+MS+SH, which combines multiscale detection and sharpening\footnote{We trained the top-performing SE+MS+SH model with $4\cdot10^6$ patches compared to $10^6$ for the other SE models, further increasing ODS by $\app.004$.}, yields an ODS of .75. In all cases OIS, which is measured using a separate optimal threshold per image, is about 2 points higher than ODS.

As these results indicate, both sharpening and multiscale detection improve accuracy. To further analyze these enhancements, we introduce a new evaluation metric: recall at 50\% precision (R50), which measures accuracy in the high recall regime. SE achieves R50 of .90 and SE+MS does not improve this. In contrast, SE+SH boosts R50 considerably to .93. This increase in recall for the SE variants can clearly be seen in \figref{fig:pr:bsds-variants}. The MS variants, on the other hand, improve precision in the low-recall regime. Overall, both SH and MS improve AP\footnote{We discovered an error in the BSDS evaluation which overestimates AP by 1\%. For consistency with past results, however, we used the code as is.}, with SE+SH+MS achieving an AP of .80.

The runtime of the four variants is reported in the last column of \tableref{table:res:bsds}. SE runs at a frame rate of 30hz, enabling real time processing. Both SH and MS slow the detector, with MS incurring a higher cost. Nevertheless, SE+SH runs at over 12hz while achieving excellent accuracy. Indeed, in the high recall regime, which is necessary for many common scenarios, SE+SH achieves top results. Given its speed and high recall, we expect SE+SH to be the default variant used in practice.

\ResultsTable{
  Human                                   & .80 & .80 &  -  & - & - \\\hline
  Canny                                   & .60 & .63 & .58 & .75 & 15 \\
  Felz-Hutt \cite{Felzenszwalb2004IJCV}   & .61 & .64 & .56 & .78 & 10 \\
  Normalized Cuts \cite{Cour2005CVPR}     & .64 & .68 & .45 & .81 & - \\
  Mean Shift \cite{Comaniciu2002PAMI}     & .64 & .68 & .56 & .79 & - \\
  Hidayat-Green \cite{Hidayat2009BMVC}    & .62$^\dagger$ & - & - & - & 20 \\
  BEL \cite{Dollar2006CVPR}               & .66$^\dagger$ & - & - & - & 1/10 \\
  Gb \cite{Leordeanu2014PAMI}             & .69 & .72 & .72 & .85 & 1/6 \\
  gPb + GPU \cite{Catanzaro2009ICCV}      & .70$^\dagger$ & - & - & - & 1/2$^\ddagger$ \\
  \hline
  ISCRA \cite{Ren2013CVPR}                & .72 & .75 & .46 & .89 & 1/30$^\ddagger$ \\
  gPb-owt-ucm \cite{Arbelaez2011PAMI}     & .73 & .76 & .73 & .89 & 1/240\\
  Sketch Tokens \cite{Lim2013CVPR}        & .73 & .75 & .78 & .91 & 1 \\
  DeepNet \cite{Kivinen2014AISTATS}       & .74 & .76 & .76 & - & 1/5$^\ddagger$ \\
  SCG \cite{Ren2012NIPS}                  & .74 & .76 & .77 & .91 & 1/280\\
  SE+multi-ucm \cite{Arbelaez2014CVPR}    & \bf{.75} & \bf{.78} & .76 & .91 & 1/15 \\
  \hline
  SE                                      & .73 & .75 & .77 & .90 & \bf{30} \\
  SE+SH                                   & .74 & .76 & .79 & \bf{.93} & 12.5 \\
  SE+MS                                   & .74 & .76 & .78 & .90 & 6 \\
  SE+MS+SH                                & \bf{.75} & .77 & \bf{.80} & \bf{.93} & 2.5 \\
}
{Results on BSDS500. \footnotesize{$^\dagger$BSDS300 results. $^\ddagger$Utilizes the GPU.}}{table:res:bsds}

\subsection{BSDS500 Results}\label{sec:results:bsds}

\begin{figure}\center
  \includegraphics[width=.42\textwidth]{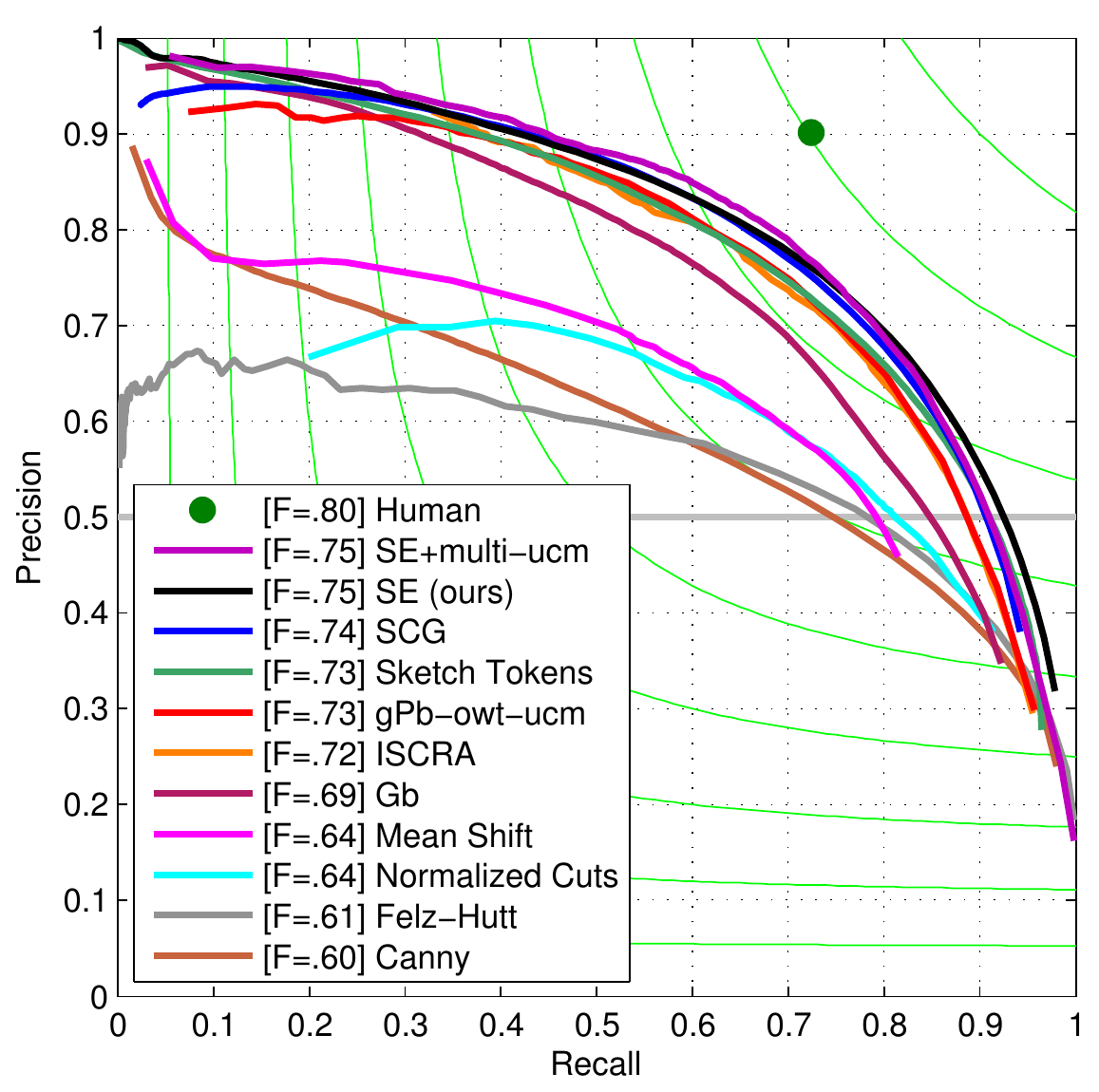}
\Caption{Results on BSDS500. Structured edges (SE) and SE coupled with hierarchical multiscale segmentation (SE+multi-ucm) \cite{Arbelaez2014CVPR} achieve top results. For the SE result we report the SE+MS+SH variant. See \tableref{table:res:bsds} for additional details including method citations and runtimes. SE is orders of magnitude faster than nearly all edge detectors with comparable accuracy.}
 \label{fig:pr:bsds}
\end{figure}

We compare our edge detector against competing methods, reporting both accuracy and runtime. Precision/recall curves are shown in \figref{fig:pr:bsds} and summary statistics are in \tableref{table:res:bsds}.

Our full approach, SE+MS+SH, outperforms all state-of-the-art approaches \cite{Ren2012NIPS,Kivinen2014AISTATS,Lim2013CVPR,Arbelaez2011PAMI,Ren2013CVPR}. We improve ODS/OIS by 1 point over competing methods and AP/R50 by 2 points. Our edge detector is particularly effective in the high recall regime. The only method with comparable accuracy is SE+multi-ucm \cite{Arbelaez2014CVPR} which couples our SE+MS detector with a hierarchical multiscale segmentation approach.

SE+MS+SH is orders of magnitude faster than nearly all edge detectors with comparable accuracy, see last column of \tableref{table:res:bsds}. All runtimes are reported on $480\times320$ images. Our approach scales linearly with image size and and is parallelized across four cores. While many competing methods are likewise linear, they have a much higher cost per pixel. The single scale variant of our detector, SE+SH, further improves speed by $5\times$ with only minor loss in accuracy and no loss in recall. Finally, SE runs at $30$hz while still achieving competitive accuracy.

In comparison to other learning-based approaches to edge detection, we considerably outperform BEL \cite{Dollar2006CVPR} which computes edges independently at each pixel given its surrounding image patch. We also outperform Sketch Tokens \cite{Lim2013CVPR} in both accuracy and runtime performance. This may be the result of Sketch Tokens using a fixed set of classes for selecting split criterion at each node, whereas our structured forests can capture finer patch edge structure. Moreover, our structured output leads to significantly smoother edge maps, see \figref{fig:res:bsds}. Finally, Kivinen \etal~\cite{Kivinen2014AISTATS} recently trained deep networks for edge detection; unfortunately, we were unable to obtain results from the authors to perform detailed comparisons.

\subsection{NYUD Results}\label{sec:results:nyu}

\begin{figure} \center
  \includegraphics[width=.42\textwidth]{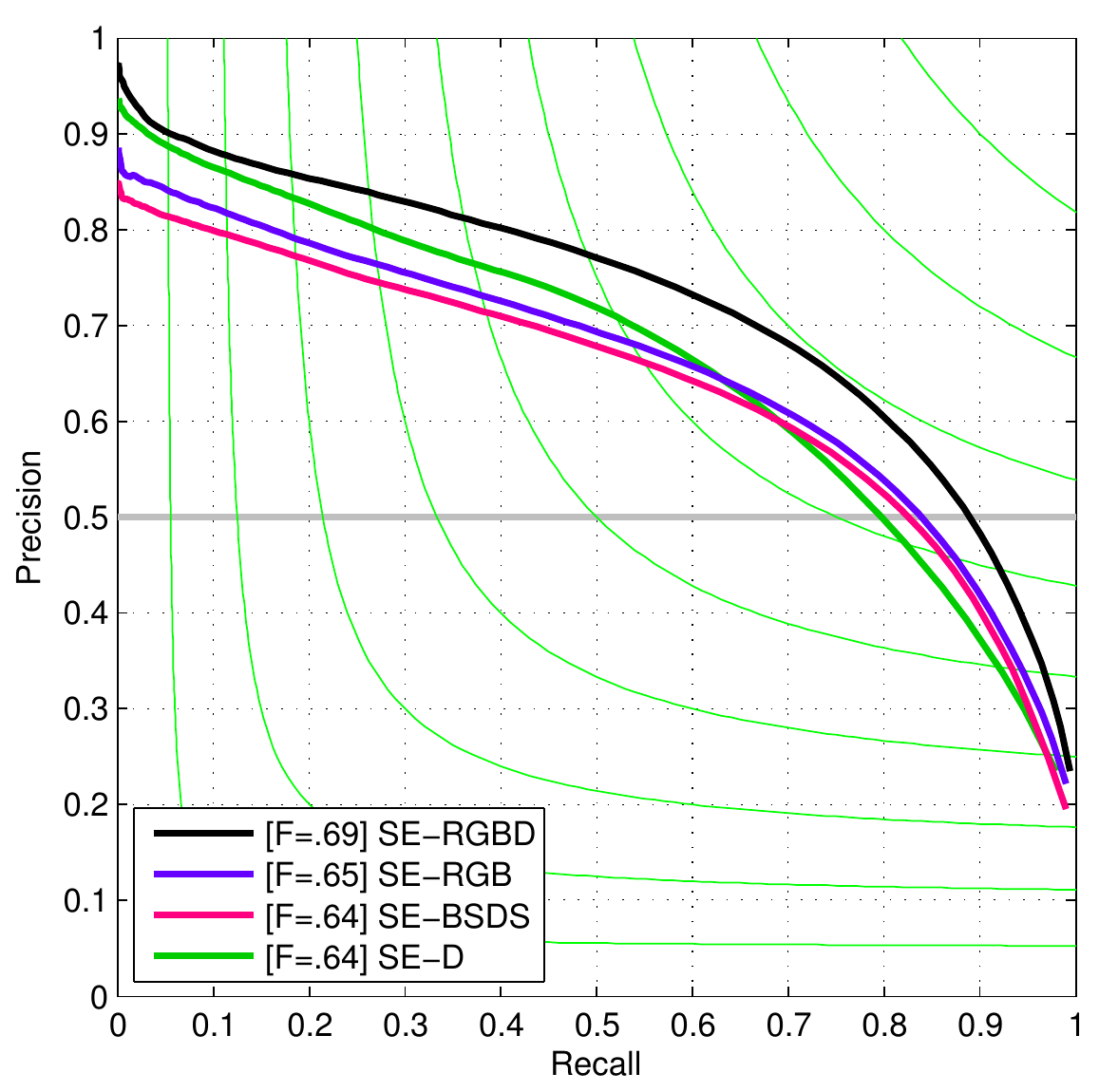}
\Caption{Precision/recall curves on NYUD using different image modalities. SE-BSDS is the RGB model trained on the BSDS dataset. See \tableref{table:res:nyu} and \ref{table:res:generalization} and text for details.}
 \label{fig:pr:nyu-variants}\vspace{3mm}
\end{figure}

\ResultsTable{
  gPb-owt-ucm \cite{Arbelaez2011PAMI}     & .63 & .66 & .56 & .79 & 1/360 \\
  Silberman \cite{Silberman2012ECCV}      & .65 & .66 & .29 & .84 & 1/360+ \\
  gPb+NG \cite{Gupta2013CVPR}             & .68 & .71 & .63 & .86 & 1/375 \\
  SE+NG+ \cite{Gupta2014ECCV}             & \bf{.71} & \bf{.72} & \bf{.74} & \bf{.90} & 1/15 \\
  \hline
  SE-D                                    & .64 & .65 & .66 & .80 & \bf{7.5} \\
  SE-RGB                                  & .65 & .67 & .65 & .84 & \bf{7.5} \\
  SE-RGBD                                 & .69 & .71 & .72 & .89 & 5 \\
}
{Results on the NYUD dataset \cite{Silberman2012ECCV}.}{table:res:nyu}

The NYU Depth (NYUD) dataset~\cite{Silberman2012ECCV} is composed of $1449$ pairs of RGB and depth images with corresponding semantic segmentations. The dataset was adopted independently for edge detection by Ren and Bo~\cite{Ren2012NIPS} and by Gupta \etal~\cite{Gupta2013CVPR}. In practice, the two dataset variants have significant differences. Specifically,~\cite{Ren2012NIPS} proposed a different train/test split and used half resolution images. Instead,~\cite{Gupta2013CVPR} use the train/test split originally proposed by Silberman \etal~\cite{Silberman2012ECCV} and full resolution images. In our previous work we used the version from~\cite{Ren2012NIPS} in our experiments; in the present work we switch to the version from Gupta \etal~\cite{Gupta2013CVPR} as more methods have been evaluated on this variant and it utilizes the full resolution images.

Gupta \etal~\cite{Gupta2013CVPR} split the NYUD dataset into 381 training, 414 validation, and 654 testing images and generated ground truth edges from the semantic segmentations provided by~\cite{Silberman2012ECCV}. The original $640\times480$ images are cropped to discard boundary regions with missing ground truth. Finally the maximum slop allowed for correct matches of edges to ground truth during evaluation is increased from .0075 of the image diagonal to .011. This is necessary to compensate for the relatively inexact localization of the ground truth.

\begin{figure*}\center
\begin{tabular}{r@{\hskip 1mm}c@{\hskip 1mm}c@{\hskip 1mm}c@{\hskip 1mm}c@{\hskip 1mm}c}
  \nyudFigHelper{}{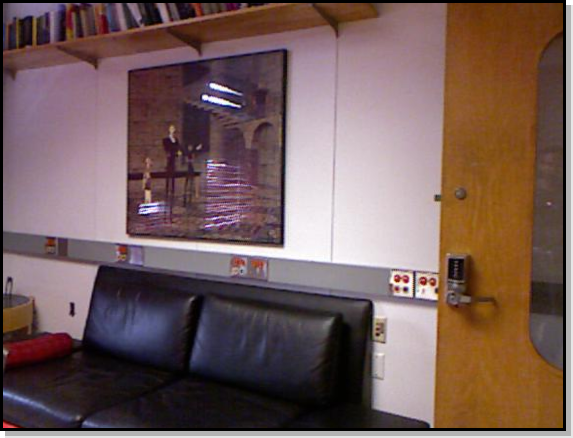}\vspace{-.9mm}\\
  \nyudFigHelper{depth}{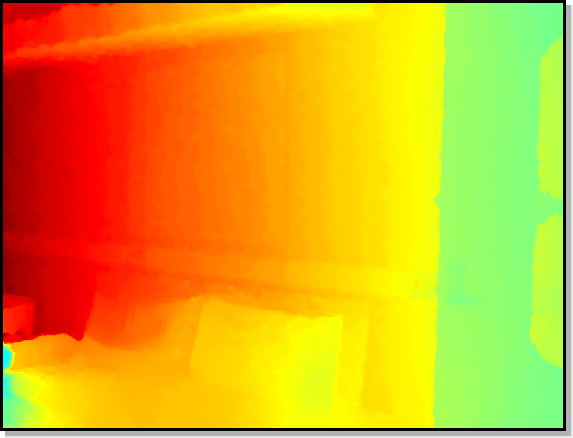}\vspace{-.9mm}\\
  \nyudFigHelper{ground truth}{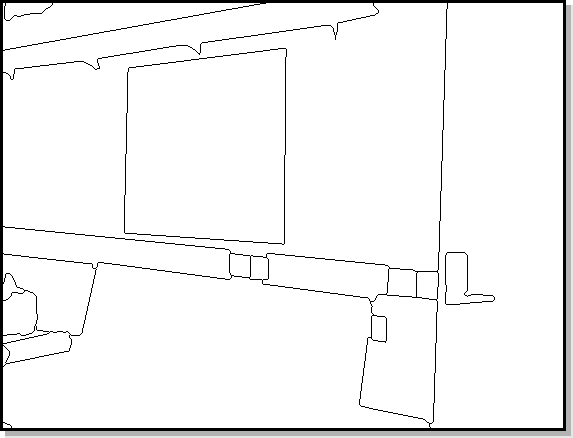}\vspace{1.5mm}\\
  \nyudFigHelper{\hspace{-25mm}SE-D}{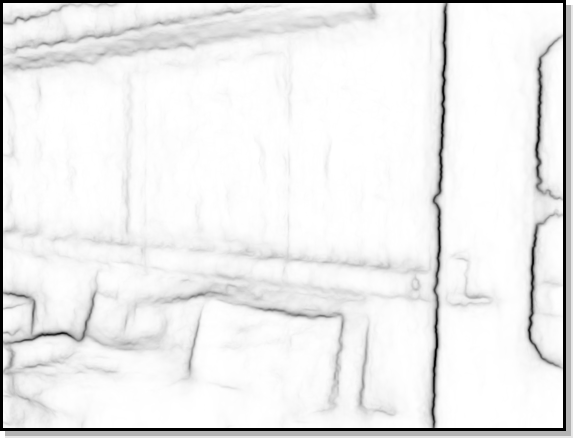}\vspace{-.9mm}\\
  \nyudFigHelper{}{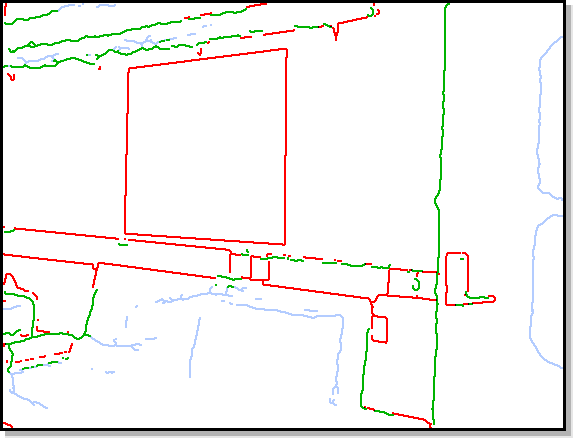}\vspace{1.5mm}\\
  \nyudFigHelper{\hspace{-25mm}SE-RGB}{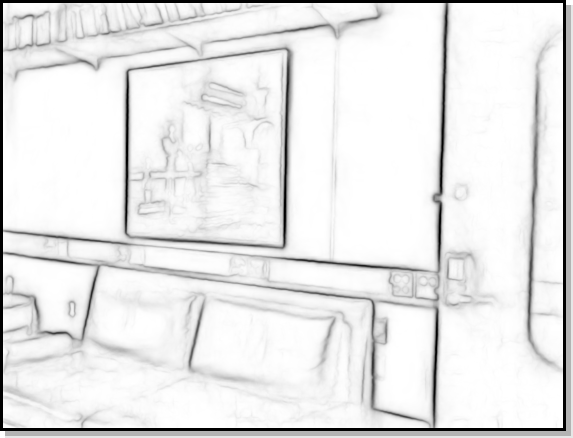}\vspace{-.9mm}\\
  \nyudFigHelper{}{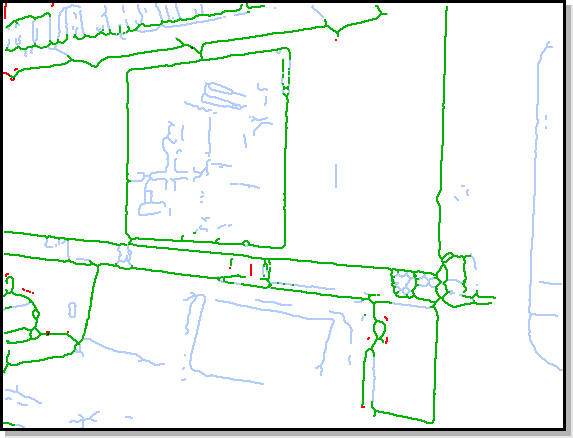}\vspace{1.5mm}\\
  \nyudFigHelper{\hspace{-25mm}SE-RGBD}{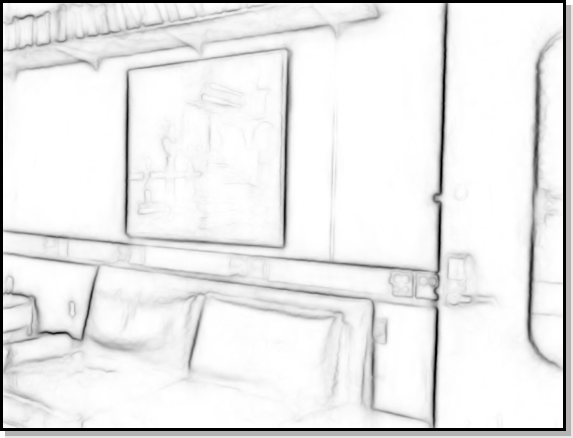}\vspace{-.9mm}\\
  \nyudFigHelper{~}{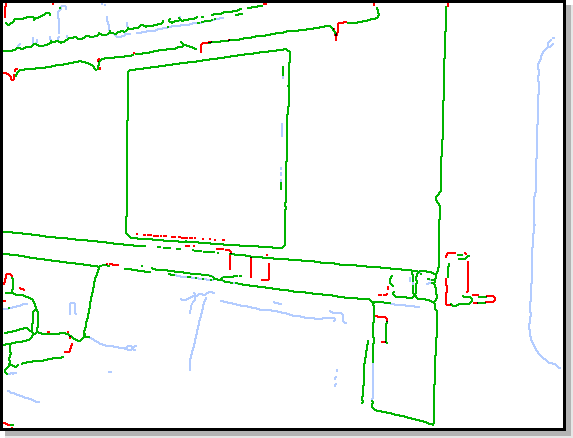}\vspace{-.9mm}\\
\end{tabular}\vspace{-1mm}
\Caption{Edge detection results on the NYUD dataset. Edges from depth features (SE-D) have good precision, edges from intensity features (SE-RGB) give better recall, and simultaneous use of intensity and depth (SE-RGBD) gives best results. For details about visualizations of matches and errors see \figref{fig:res:bsdseval}; all visualizations were generated using each detector's ODS threshold.}
\label{fig:res:nyu}\vspace{-2mm}
\end{figure*}

Example SE results are shown in \figref{fig:res:nyu}. We treat the depth channel in the same manner as the other color channels. Specifically, we recompute the gradient channels over the depth channel (with identical parameters) resulting in 11 additional channels. Precision/recall curves for SE+SH with different image modalities are shown in \figref{fig:pr:nyu-variants}. Use of depth information only (SE-D) gives good precision as strong depth discontinuities nearly always correspond to edges. Use of intensity information only (SE-RGB) gives better recall as nearly all edges have intensity discontinuities but not all edges have depth discontinuities. As expected, simultaneous use of intensity and depth (SE-RGBD) substantially improves results.

Summary statistics are given in \tableref{table:res:nyu}. Runtime is slower than on BSDS as NYUD images are higher resolution and features must be computed over both intensity and depth. For these results we utilized the SE+SH variant which slightly outperformed SE+MS+SH on this dataset.

\begin{figure} \center
  \includegraphics[width=.42\textwidth]{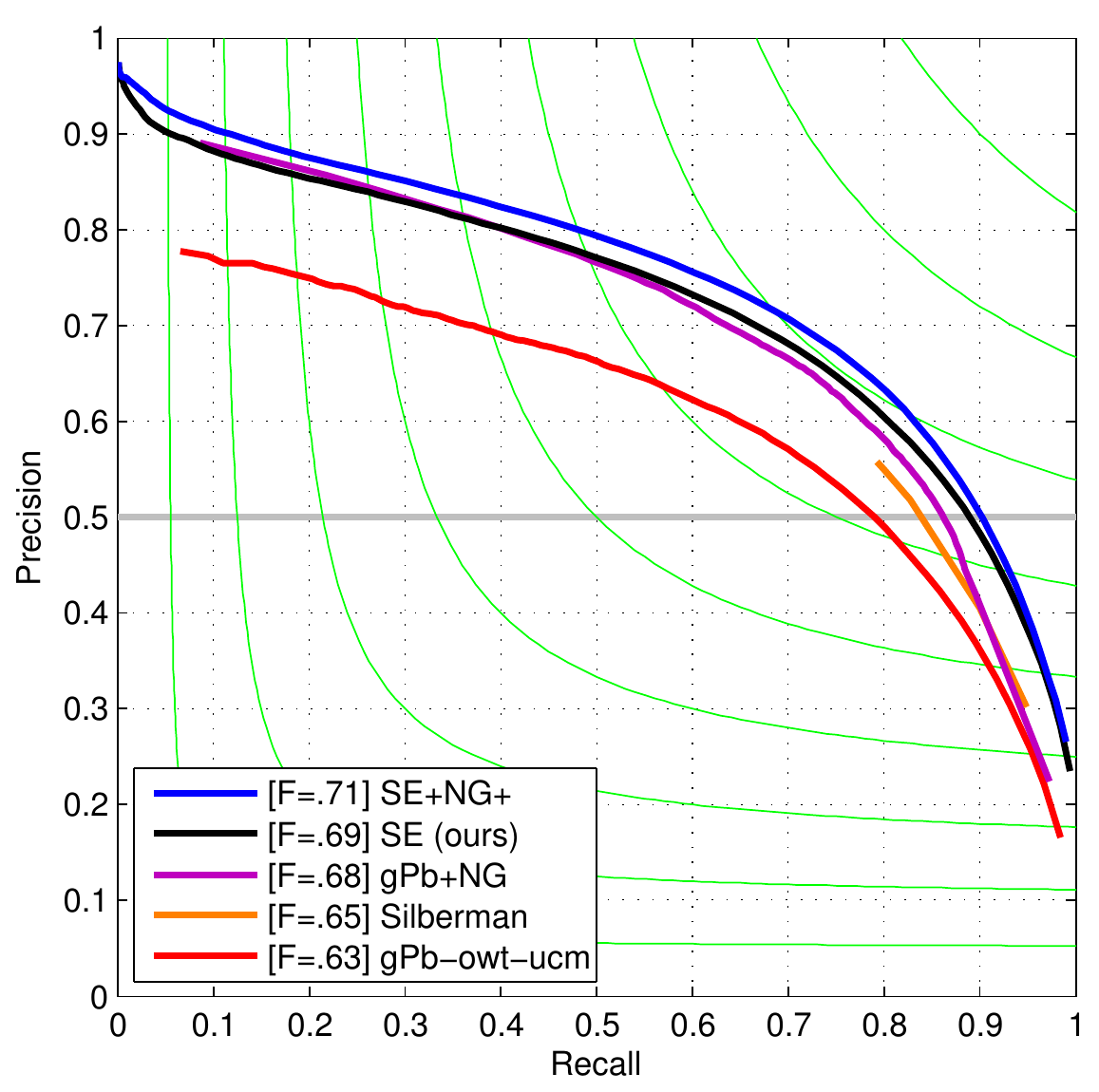}
\Caption{Results on NYUD. Structured edges (SE) and SE coupled with depth normal gradient (SE+NG+) \cite{Gupta2013CVPR,Gupta2014ECCV} achieve top results. For the SE result we report the SE+SH variant. See \tableref{table:res:nyu} for additional details including citations and runtimes.}
 \label{fig:pr:nyu}
\end{figure}

In \tableref{table:res:nyu} and \figref{fig:pr:nyu} we compare our approach, SE+SH, to a number of state-of-the-art approaches, including gPb-owt-ucm (color only), Silberman \etal's RGBD segmentation algorithm~\cite{Silberman2012ECCV}, and detectors from Gupta \etal~\cite{Gupta2013CVPR,Gupta2014ECCV} that explicitly estimate depth normal gradient (NG). While our approach naively utilizes depth (treating the depth image identically to the intensity image), we outperform nearly all competing methods, including gPb+NG~\cite{Gupta2013CVPR}. Gupta \etal~\cite{Gupta2014ECCV} obtain top results by coupling our structured edges detector with depth normal gradients and additional cues (SE+NG+). Finally, for a comparison of SE to Ren and Bo's SCG~\cite{Ren2012NIPS} on the alternate version of NYUD we refer readers to our previous work \cite{Dollar2013ICCV} (in the alternate setup SE likewise outperforms SCG across all modalities while retaining its speed advantage).

\subsection{Cross dataset generalization}\label{sec:results:cross}

To study the ability of our approach to generalize across datasets we ran a final set of experiments. In \tableref{table:res:generalization} we show results on NYUD using structured forests trained on BSDS and also results on BSDS using structured forests trained on NYUD. For these experiments we use intensity images only. Note that images in the BSDS and NYUD datasets are qualitatively quite different, see \figref{fig:res:bsds} and \ref{fig:res:nyu}, respectively.

\tableref{table:res:generalization}, top, compares results on NYUD of the NYUD and BSDS trained models. Across all performance measure, scores degrade by about 1 point when using the BSDS dataset for training. Precision/recall curves on NYUD for the NYUD model (SE-RGB) and the BSDS model (SE-BSDS) are shown in \figref{fig:pr:nyu-variants}. The resulting curves align closely. The minor performance change is surprising given the different statistics of the datasets. Results on BSDS of the BSDS and NYUD models, shown in \tableref{table:res:generalization}, bottom, are likewise similar.

These experiments provide strong evidence that our approach could serve as a general purpose edge detector without the necessity of retraining. We expect this to be a critical aspect of our detector allowing for its widespread applicability.

\ResultsTable{
  NYUD / NYUD                             & .65 & .67 & .65 & .84 & 7.5 \\
  BSDS / NYUD                             & .64 & .66 & .63 & .83 & 7.5 \\\hline
  BSDS / BSDS                             & .75 & .77 & .80 & .93 & 2.5 \\
  NYUD / BSDS                             & .73 & .74 & .77 & .91 & 2.5 \\
}
{Cross-dataset generalization for Structured Edges. TRAIN/TEST indicates the training/testing dataset used. }{table:res:generalization}

\section{Discussion}\label{sec:discussion}

Our approach is capable of realtime frame rates while achieving state-of-the-art accuracy. This may enable new applications that require high-quality edge detection and efficiency. For instance, our approach may be well suited for video segmentation or for time sensitive object recognition tasks such as pedestrian detection.

Our approach to learning structured decision trees may be applied to a variety of problems. The fast and direct inference procedure is ideal for applications requiring computational efficiency. Given that many vision applications contain structured data, there is significant potential for structured forests in other applications.

In conclusion, we propose a structured learning approach to edge detection. We describe a general purpose method for learning structured random decision forest that robustly uses structured labels to select splits in the trees. We demonstrate state-of-the-art accuracies on two edge detection datasets, while being orders of magnitude faster than most competing state-of-the-art methods.

Source code is available online.

{\bibliographystyle{ieee}\bibliography{biblio}}

\end{document}

%% file: preamble.tex
\usepackage{times,graphicx,amsmath,amssymb,arydshln,booktabs,xspace}
\usepackage[pagebackref=true,colorlinks=true,bookmarks=false]{hyperref}

\let\mypdfximage\pdfximage\def\pdfximage{\immediate\mypdfximage}

\makeatletter
\DeclareRobustCommand\onedot{\futurelet\@let@token\@onedot}
\def\@onedot{\ifx\@let@token.\else.\null\fi\xspace}
\def\eg{e.g\onedot} \def\Eg{E.g\onedot}
\def\ie{i.e\onedot} 
 
\def\etc{etc\onedot} 
 
\def\etal{et~al\onedot} 
  
\makeatother

\DeclareRobustCommand{\figref}[1]{Figure~\ref{#1}}
\DeclareRobustCommand{\figsref}[1]{Figures~\ref{#1}}
\DeclareRobustCommand{\secref}[1]{\S\ref{#1}}
\DeclareRobustCommand{\tableref}[1]{Table~\ref{#1}}
\DeclareRobustCommand{\eqnref}[1]{Eqn.~(\ref{#1})}

\newcommand{\eqn}[2]{{#1\begin{eqnarray}#2\end{eqnarray}}}
\newcommand{\eqns}[2]{{#1\begin{eqnarray*}#2\end{eqnarray*}}}
\newcommand{\app}{\raise.17ex\hbox{$\scriptstyle\sim$}}
\newcommand{\Caption}[1]{\caption{\small#1}}
\newcommand{\hide}[1]{}

\def\R{\mathbb{R}}
\def\X{\mathcal{X}}
\def\Y{\mathcal{Y}}
\def\Z{\mathcal{Z}}
\def\XY{\mathcal{S}}
\def\C{\mathcal{C}}
\def\map{\Pi}

\newcommand{\ResultsTable}[3]{
  \renewcommand{\arraystretch}{1.3}\def\cp{\hspace{4mm}}
  \begin{table}\centering\small
    \begin{tabular}{@{}l@{\cp}|c@{\cp}c@{\cp}c@{\cp}c@{\cp}|c@{\cp}}
      & ODS & OIS & AP & R50 & FPS \\\toprule[0.5mm]
      #1
    \end{tabular}\vspace{2mm}
  \caption{#2 }
  \label{#3}\end{table}
}

\newcommand{\bsdsFigHelper}[2]{
  \hspace{-2mm}\rotatebox{90}{\begin{minipage}{
    .125\linewidth}\center#1\end{minipage}} &
  \includegraphics[width=.187\linewidth]{figures/results/157087#2} &
  \includegraphics[width=.187\linewidth]{figures/results/97010#2} &
  \includegraphics[width=.187\linewidth]{figures/results/196062#2} &
  \includegraphics[width=.187\linewidth]{figures/results/69000#2} &
  \includegraphics[width=.187\linewidth]{figures/results/81066#2}\vspace{-.25mm}\\
}

\newcommand{\nyudFigHelper}[2]{
  \hspace{-2mm}\rotatebox{90}{\begin{minipage}{
    .13\linewidth}\center#1\end{minipage}} &
  \includegraphics[width=.175\linewidth]{figures/results/img_5043#2} &
  \includegraphics[width=.175\linewidth]{figures/results/img_6433#2} &
  \includegraphics[width=.175\linewidth]{figures/results/img_6279#2} &
  \includegraphics[width=.175\linewidth]{figures/results/img_6002#2} &
  \includegraphics[width=.175\linewidth]{figures/results/img_5328#2}
}

\newcommand{\incp}[1]{\includegraphics[width=.465\linewidth]
 {figures/sweepsBsds/#1.pdf}\vspace{-.2mm}}